\PassOptionsToPackage{hmargin=1.25cm, vmargin=2cm}{geometry}
\documentclass[3p, twocolumn]{elsarticle}
\usepackage[hidelinks]{hyperref}
\usepackage{lineno}
\usepackage{array}
\newcolumntype{L}[1]{>{\centering\arraybackslash}m{#1}}
\usepackage{multirow}
\usepackage{subfigure}
\usepackage{xcolor}
\usepackage{tikz,stackengine}
\usepackage[printwatermark]{xwatermark}

\def\eg{\emph{e.g}. } 
\def\ie{\emph{i.e}. }

 \def\etal{\emph{et al}. }

\newwatermark*[allpages,color=red!50,scale=3,xpos=-8cm,ypos=-12cm]{	\includegraphics[width=0.07\linewidth]{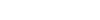}}
\begin{document}

\begin{frontmatter}
\title{Mobile Touchless Fingerprint Recognition:\\ Implementation, Performance and Usability Aspects}

\author[HDA,FUB]{Jannis Priesnitz\corref{mycorrespondingauthor}}
\cortext[mycorrespondingauthor]{Corresponding author}
\ead{jannis.priesnitz@h-da.de}
\ead[url]{dasec.h-da.de}
\author[UCS]{Rolf Huesmann}
\author[HDA]{Christian Rathgeb}
\author[FUB]{Nicolas Buchmann}
\author[HDA]{Christoph Busch}

\address[HDA]{da/sec -- Biometrics and Internet
	Security Research Group,
	Hochschule Darmstadt,
	Sch\"offerstraße 8b, 64295
	Darmstadt, Germany
}

\address[UCS]{UCS -- User-Centered Security Research Group,
	Hochschule Darmstadt,
	Sch\"offerstraße 8b, 64295
	Darmstadt, Germany
}

\address[FUB]{Freie Universit\"at Berlin, Takustraße 9, 14195 Berlin, Germany}

\begin{abstract}
This work presents an automated touchless fingerprint recognition system for smartphones. We provide a comprehensive description of the entire recognition pipeline and discuss important requirements for a fully automated capturing system. Also, our implementation is made publicly available for research purposes. During a database acquisition, a total number of 1,360 touchless and touch-based samples of 29 subjects are captured in two different environmental situations. Experiments on the acquired database show a comparable performance of our touchless scheme and the touch-based baseline scheme under constrained environmental influences. A comparative usability study on both capturing device types indicates that the majority of subjects prefer the touchless capturing method. Based on our experimental results we analyze the impact of the current COVID-19 pandemic on fingerprint recognition systems. Finally, implementation aspects of touchless fingerprint recognition are summarized. 

\end{abstract}


\end{frontmatter}
\section{Introduction}
Fingerprints are one of the most important biometric characteristic due to their known uniqueness and persistence properties. Fingerprint recognition systems are not only used worldwide by law enforcement and forensic agencies, they are also deployed in the mobile devices as well as in nation-wide applications. 
The vast majority of fingerprint capturing schemes requires contact between the finger and the capturing device's surface. These systems suffer from distinct problems, e.g. low contrast caused by dirt or humidity on the capturing device plate or latent fingerprints of previous users (ghost fingerprints). Especially in multi-user applications hygienic concerns lower
the acceptability of touch-based fingerprint systems and hence, limit their deployment. In a comprehensive study, Okereafor et al. \cite{okereafor_fingerprint_2020} analyze the risk of an infection by touch-based fingerprint recognition schemes and the hygienic concerns of their users. The authors conclude that touch-based fingerprint recognition carries a high risk of an infection if a previous user has contaminated the capturing device surface, e.g. with the SARS-CoV-2 virus.

\begin{table*}[t]
	\caption{Overview on selected recognition workflows with biometric performance. (Device type: P = Prototypical hardware, S = Smartphone,  W = Webcam) }
	\label{table_performance}
	\footnotesize
	\centering
	\begin{tabular}{|L{0.165\linewidth}| L{0.04\linewidth}|L{0.05\linewidth}|L{0.08\linewidth}|L{0.07\linewidth}|L{0.08\linewidth}| L{0.09\linewidth}|L{0.0825\linewidth}|L{0.125\linewidth}|}\hline
		\textbf{\rotatebox{0}{Authors} } & 
		\textbf{\rotatebox{0}{Year}} & 
		\textbf{\rotatebox{0}{\shortstack{Device \\ type }}} &
		\textbf{\rotatebox{0}{\shortstack{Mobile / \\Stationary$\:$}}} &
		\textbf{\rotatebox{0}{\shortstack{Amount \\ of fingers$\:$}}} & 
		\textbf{\rotatebox{0}{\shortstack{Automatic$\:$ \\ capturing$\:$}}} & 
		\textbf{\rotatebox{0}{\shortstack{Free finger \\ positioning$\:$}}} & 
		\textbf{\rotatebox{0}{\shortstack{On-device \\processing$\:$}}} &
		\textbf{Biometric Performance}
		\\\hline
		Hiew et al. \cite{hiew2007digital} & 
		2007 &
		P & S & 1 & N & N & N & EER: 5.63\%
		\\\hline
		Piuri et al. \cite{piuri2008fingerprint} & 
		2008 &
		W & S & 1 & N & N & N & EER: 0.04\%
		\\\hline
		Wang et al. \cite{wang2009novel} &
		2009 &
		P & S &  1 & N & N &  N & EER: 0.0\% \footnote{}
		\\\hline
		Kumar and Zhou \cite{kumar2011contactless} & 
		2011 &
		W & S & 1 & N & N & N & EER: 0.32\%
		\\\hline
		Derawi et al. \cite{derawi2011fingerprint} & 
		2011 &
		S & S & 1 & N & N & N & EER: 4.5\%
		\\\hline
		Noh et al. \cite{noh2011touchless} &
		2011 &
		P & S & 5 & Y & N & N & EER: 6.5\%
		\\\hline
		Stein et al. \cite{stein2013video-based} & 
		2013 & S & M & 1 & Y & Y & Y & EER: 1.2\%
		\\\hline
		Raghavendra et al. \cite{raghavendra2014low-cost} &
		2014 &
		P & S & 1 & N & Y & N & EER: 6.63\% 
		\\\hline
		Tiwari et al. \cite{tiwari2015touch-less}&
		2015 &
		S & M & 1 & N & Y & N & EER: 3.33\% 
		\\\hline
		Sankaran et al. \cite{sankaran2015smartphone} &
		2015 &
		S & M & 1 & Y & N & N & EER: 3.56\%
		\\\hline
		Carney et al. \cite{carney2017multi-finger} &
		2017 &
		S& M & 4 & Y & N & Y & FAR = 0.01\% $@$ FRR = 1\%
		\\\hline
		Deb et al. \cite{deb2018matching} & 		
		2018 & 
		S & M & 1 & Y & Y & Y & TAR = 98.55\% $@$ FAR = 1.0\%	
		\\\hline
		Weissenfeld et al. \cite{weissenfeld2018contactless} & 
		2018 &
		P & M & 4 & Y & Y & Y & FRR 1.2\% $@$ FAR 0.01\% %
		\\\hline
		Birajadar et al. \cite{birajadar_towards_2019} & 
		2019 &
		S & M & 1 & Y & N & N & EER: 1.18\% 
		\\\hline
		Our method &
		2021 &
		S &	M &	4 & Y & Y & Y & EER: 5.36\%
		\\\hline
	\end{tabular}
\end{table*}
\begin{figure*}[t]
	\centering
	\includegraphics[width=0.999\linewidth]{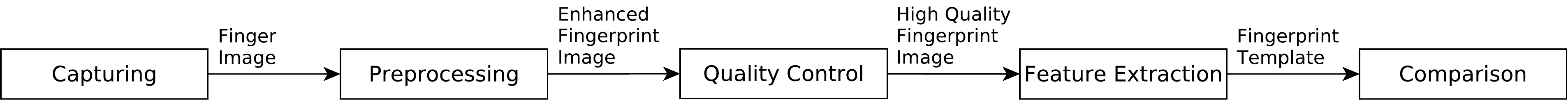}
	\caption{Overview on the most relevant steps of our proposed method.}
	\label{fig:generaloverviewposterflat}
\end{figure*}
To tackle these shortcomings of touch-based schemes, touchless fingerprint recognition systems have been researched for more than a decade. Touchless capturing schemes operate without any contact between the finger and the capturing device.
Several contributions to the research area paved the way for a practical implementation of touchless capturing schemes. Specialized stationary capturing devices based on multi-camera setups combined with a powerful processing have already been implemented in a practical way \cite{chen20063d}. However, to the best of the authors' knowledge no approach to mobile touchless recognition scheme based on of-the-shelf components has been documented so far. 

In this work, we propose a mobile touchless fingerprint recognition scheme based for smartphones. We provide a fully automated capturing scheme with integrated quality assessment in form of an Android app. All components of the processing pipeline will be made publicly available for research purposes.
The rest of the recognition pipeline is composed of open source algorithms for feature extraction and comparison. 

To benchmark our proposed recognition pipeline we acquired a database under real life conditions. A number of 29 subjects have been captured by two touchless capturing devices in different environmental situations. Moreover, touch-based samples have been acquired as baseline and to measure the interoperability between both schemes. 
A usability study, which has been conducted after the capturing, reviews the users' experiences with both capturing device types. 
This paper details on every implementation step of the processing pipeline and  summarizes our database acquisition setup in detail. Further, we evaluate the biometric performance of our acquired database and discusses the outcomes of our usability study.
Based on our experimental results we elaborate on the impact of the current COVID-19 pandemic on fingerprint recognition  in terms of biometric performance and user acceptance.
Further we summarize implementation aspects which we consider as beneficial for mobile touchless fingerprint recognition. 

The rest of the paper is structured as follows: 
Section~\ref{sec:related_work} gives an overview on touchless end-to-end schemes proposed in the scientific literature. 
In Section~\ref{sec:recognition_pipeline}, the proposed processing pipeline is presented. 
In Section~\ref{sec:experimental_setup}, we describe our experimental setup and provide details about the captured database and the usability study. 
The results of our experiments are reported in Section~\ref{sec:results}. The influence of the COVID-19 pandemic on fingerprint recognition is discussed in Section~\ref{sec:covid_impact}. 
Section~\ref{sec:imeplmentation_aspects} discusses implementation aspects. 
Finally, Section~\ref{sec:conclusion} concludes.

\footnotetext{The authors only report the EER on a score level fusion of 8 sequentially acquired fingerprints.}

\section{Related Work}
\label{sec:related_work}
In this section, we present an overview on touchless fingerprint end-to-end solutions along with reported biometric performance. Table~\ref{table_performance} summarizes most relevant related works and their properties. For a comprehensive overview on the topic of touchless fingerprint recognition the reader is referred to \cite{priesnitz2021overview}.

The research on touchless fingerprint recognition has evolved from bulky single finger devices to more convenient multi-finger capturing schemes. First end-to-end approaches with prototypical hardware setups were presented by Hiew et al. \cite{hiew2007digital} and Wang et al. \cite{wang2009novel}. Both works employed huge capturing devices for one single finger acquisition within a hole-like guidance. 

For remote user authentication, Piuri et al. \cite{piuri2008fingerprint} and Kumar et al. \cite{kumar2011contactless} investigated the use of webcams as fingerprint capturing device. Both schemes showed a very low EER in experimental results. However, the database capturing process was not reported precisely. In addition, the usability  and user acceptance of such an approach should be further investigated. 

More recent works use smartphones for touchless fingerprint capturing. Here, a finger image is taken by a photo app and manually transferred to a remote device where the processing is performed \cite{tiwari2015touch-less, sankaran2015smartphone}. 
The improvement of the camera and processing power in current smartphones has made it possible to capture multiple fingers in in a single capture attempt and process them on the device. Stein et al. \cite{stein2013video-based} showed that it is feasible for the automated capturing of a single finger image using a smartphone. Carney et al. \cite{carney2017multi-finger} presented the first four-finger capturing scheme. Weissenfeld et al. \cite{weissenfeld2018contactless} proposed a system with a free positioning of four fingers in a mobile prototypical hardware setup. 

In summary, it can  be observed that the evolution of touchless fingerprint technologies has moved towards out-of-the-box device and a more convenient and practically relevant recognition process. 
Table~\ref{table_performance} also indicates that both an accurate recognition and a convenient capturing are hard to achieve. It should be further noted that the level of constraints during the capturing has a major influence on the recognition performance.

\begin{figure*}
	\centering
	\includegraphics[width=0.79\linewidth]{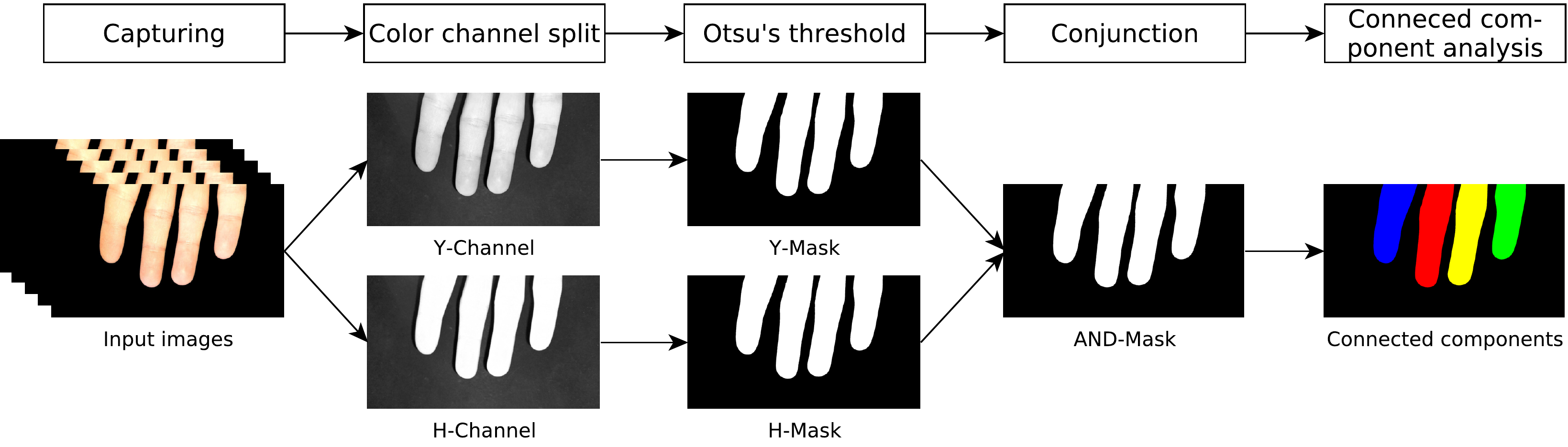}
	\caption{Overview of the segmentation of connected components from a continuous stream of input images.}
	\label{fig:segmentation}
\end{figure*}
\begin{figure*}[!t]
	\centering
	\includegraphics[width=0.9\linewidth]{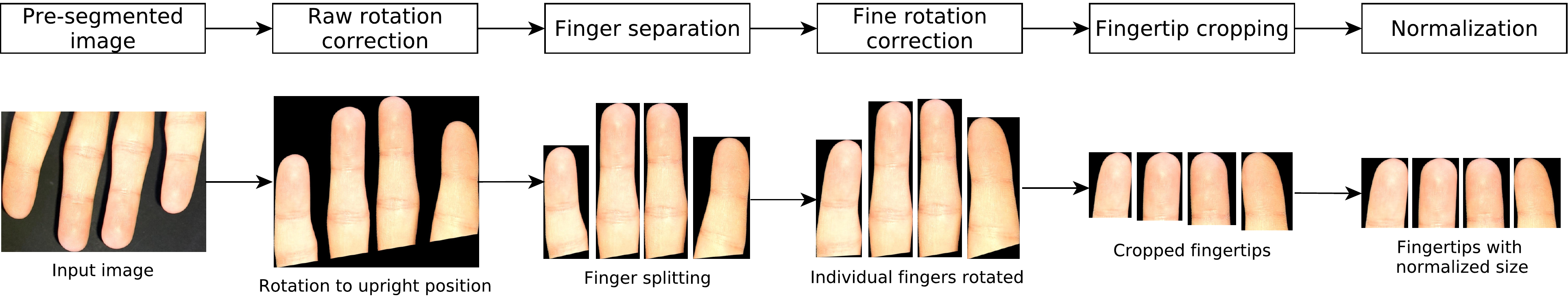}
	\caption{Overview of the coarse rotation correction, separation of fingerprint images from each other, fine rotation correction, fingertip cropping and normalization of the fingerprint size.}
	\label{fig:rot_crop_norm}
\end{figure*}
\section{Mobile Touchless Recognition Pipeline}
\label{sec:recognition_pipeline}
An unconstrained and automated touchless fingerprint recognition system  usually requires a more elaborated processing compared to touch-based schemes. Figure~\ref{fig:generaloverviewposterflat} gives an overview on the key processing steps of the proposed recognition pipeline.  
Our method features an on-device capturing, pre-processing, and quality assessment whereas the biometric identification workflow is implemented on a back-end system. This section describes each component of the recognition pipeline in detail.

The proposed method combines three implementation aspects seen as beneficial for an efficient and convenient recognition: 
\begin{itemize}
	\item A smartphone as hardware platform which captures the finger image and obtains a fingerprint images from it. 
	\item A free positioning of the fingers without guidelines or framing. 
	\item A fully automated capturing and processing with an integrated quality assessment and user feedback.	
\end{itemize}

The whole capturing and processing pipeline discussed in this section is made publicly available\footnote{Source code will be made available at: \url{https://gitlab.com/jannispriesnitz/mtfr}}. 

\subsection{Capturing}
The vast majority of mobile touchless recognition schemes rely on state-of-the-art smartphones as capturing devices. Smartphones offer a high-resolution camera unit, a powerful processor, an integrated user feedback via display and speaker, as well as a mobile internet connection for on-demand comparison against centrally stored databases. %

In our case, the capturing as well as the processing is embedded in an Android App. 
Once the recognition process is started the application analyzes the live-view image and automatically captures a finger image if the quality parameters fit the requirements. During capturing the user is able to see his/her fingers through a live-view on the screen and is able to adjust the finger position. In addition, the capturing progress is displayed.

\begin{figure*}[!t]
	\centering
	\includegraphics[width=0.85\linewidth]{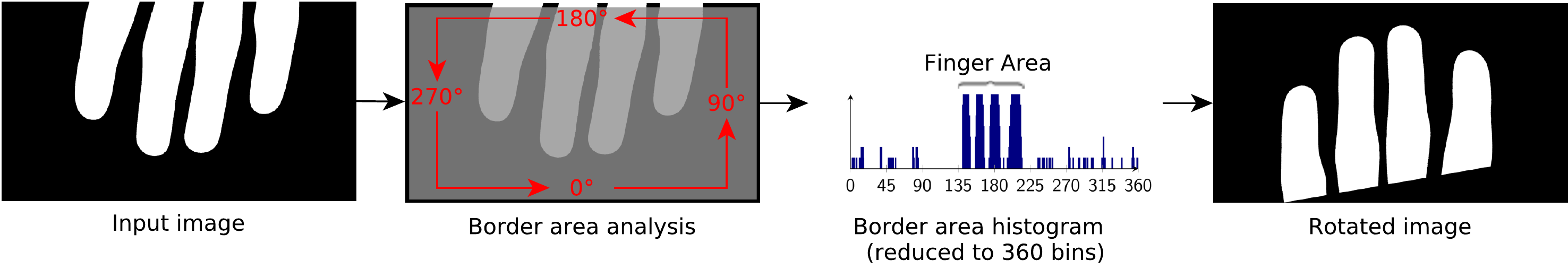}
	\caption{Detailed workflow of the coarse rotation correction.}
	\label{fig:rotationcorrection}
\end{figure*}

\begin{figure*}[!t]
	\centering
	\includegraphics[width=0.56\linewidth]{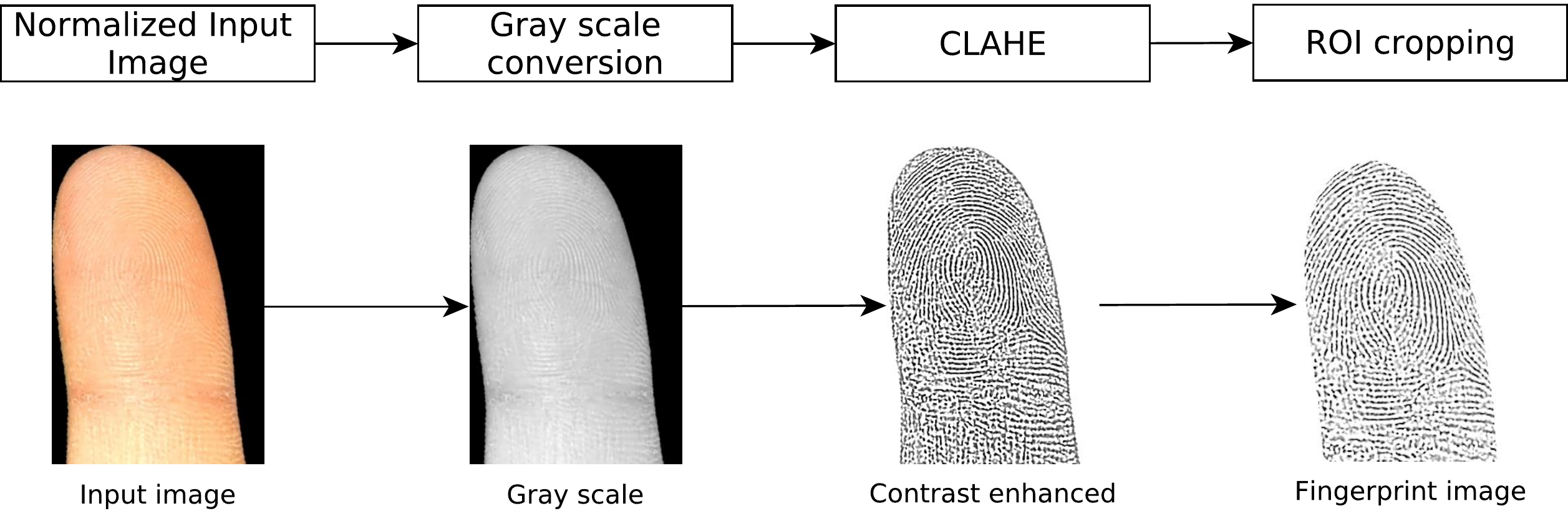}
	\caption{Overview of gray-scale conversion, application of CLAHE and cropping of the fingerprint Region Of Interest (ROI). This process is executed on every separated finger.}
	\label{fig:clahe_roi}
\end{figure*}
\subsection{Segmentation of the Hand Area}
Proposed strategies for the segmentation mainly rely on color and contrast. Many works use color models for segmenting the hand color from the background. Here an Otsu's adaptive threshold is preferable over static thresholding. Combinations of different color channels also show superior results compared to schemes based on one channel \cite{hiew2006automatic, sisodia2017conglomerate, wang2016preprocessing, malhotra_fingerphoto_2017}. 

Figure~\ref{fig:segmentation} presents an overview of the segmentation workflow. 
We adopt this method and analyze the Cr component of the yCbCr color model and the H component of the HSV color model. As first step we normalize the color channels to the full range of the histogram. Subsequently, the Otsu's threshold determines the local minimum in the histogram curve. A binary mask is created where all pixel values below the threshold are set to black and all pixels above the threshold are set to white. 

Additionally, our algorithm analyzes the largest connected components within the segmentation mask. Ideally the segmentation mask should only contain one to four dominant components: from one hand area up to four finger areas, respectively. The method also checks the size and shape of segmented areas. 

\subsection{Rotation Correction, Fingertip Detection and Normalization}

The rotation correction transforms every finger image in a way that the final fingerprint image is oriented in an upright position. 

Figure~\ref{fig:rot_crop_norm} presents an overview of the rotation correction, fingertip detection and normalization. 
Our method features two rotation steps: First a coarse rotation on the full hand and second a fine rotation on the separated finger. A robust separation and identification of the fingers requires that the hand is rotated to an upright position. Here the image border of the binary segmentation mask is analyzed. Many white border pixels indicate that the hand is placed into the sensor area from this particular direction.
For this reason, we search for the border area with the most white pixels and calculate a rotation angle from this coordinate. Figure~\ref{fig:rotationcorrection} illustrates this method. 

After the coarse rotation the fingertips are separated. To this end, the amount of contours of considerable size is compared to a pre-configured value. If there are fewer contours than expected it is most likely that the finger images contain part of the palm of the hand. In this case pixels are cut out from the bottom of the image and the sample is tested again. 
In the case of more considerable contours, the finger image is discarded in order to avoid processing wrong finger-IDs. 

An upright rotated hand area does not necessarily mean that the fingers are accurately rotated because fingers can be spread. A fine rotation is computed on every finger image to correct such cases. Here, a rotated minimal rectangle is placed around every dominant contour. This minimal rectangle is then rotated into an upright position.  

Additionally, the hight of each finger image needs to be reduced to the area which contains the fingerprint impression. Other works have proposed algorithms which search for the first finger knuckle \cite{raghavendra2013scaling-robust, stein2012fingerphoto}. We implemented a simpler method which cuts the hight of the finger image to the double of its width. In our use case this method leads to slightly less accurate result but is much more robust against outliers.

Touchless fingerprint images captured in different sessions or processed by different workflows do not necessarily have the same size. The distance between sensor and finger defines the scale of the resulting image. For a minutiae-based comparison it is crucial that both samples have the same size. Moreover, in a touchless-to-touch-based interoperability scenario the sample size has to be aligned to the standardized resolution, e.g. 500dpi. Therefore, we normalize the fingerprint image to a width of 300px. 
This size approximately corresponds to the size touch-based fingerprints captured with 500dpi. 

Together with the information which hand is captured, an accurate rotation correction also enables a robust identification of the finger ID, e.g. index, middle, ring or pinky finger. 

\subsection{Fingerprint Processing}
The pre-processed fingerprint image is aligned to resemble the impression of a touch-based fingerprint. Figure~\ref{fig:clahe_roi} presents the conversion from a finger image to a touchless fingerprint. We use the Contrast Limited Adaptive Histogram Equalization (CLAHE) on a gray scale converted fingerprint image to emphasize the ridge-line characteristics. 

Preliminary experiments showed that the used feature extractor detects many false minutiae at the border region of touchless fingerprint samples. For this reason, we crop approx. 15 pixel of the border region. That is, the segmentation mask is dilated in order to reduce the size of the fingerprint image. 

\begin{figure*}[!t]
	\centering
	\subfigure[Box setup]{\includegraphics[width=0.32\linewidth, trim=0px 0px 15px 0px, clip]{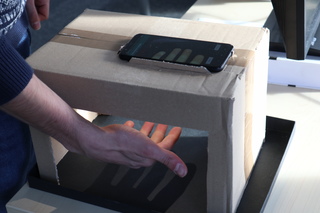}}\hfil
	\subfigure[Tripod setup]{\includegraphics[width=0.32\linewidth, trim=0px 0px 15px 0px, clip]{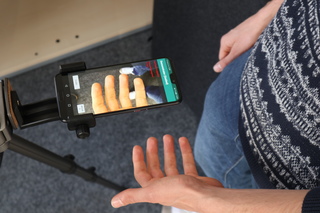}}\hfil
	\subfigure[Touch-based capturing device]{\includegraphics[width=0.32\linewidth, trim=0px 0px 15px 0px, clip]{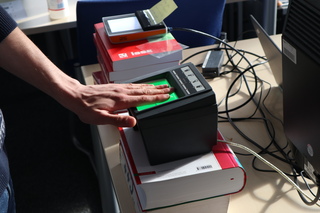}}\\
	\caption{Capturing device setups during our experiments. }\label{fig:capturing devices}
\end{figure*}

\begin{table*}[t]
	\caption{Overview on selected recognition workflows with biometric performance.}
	\label{tab:setup}
	\footnotesize
	\centering
	\begin{tabular}{|c|c|c|c|c|c|}
		\hline
		\textbf{Type} & \textbf{Setup} & \textbf{Device} & \textbf{Subjects captured} & \textbf{Rounds} & \textbf{Samples} \\
		\hline
		Touchless & Box & Google Pixel 4 & 28 & 2 & 448 \\
		\hline
		Touchless & Tripod & Huawei P20 Pro & 28 & 2 & 448 \\
		\hline
		Touch-based & -- & Crossmatch  Guardian 100 & 29 & 2 & 464 \\
		\hline
	\end{tabular}
\end{table*}
\subsection{Quality Assessment}
Quality assessment is a crucial task for touch-based and touchless fingerprint recognition schemes. We distinguish between two types of quality assessment: An integrated plausibility check at certain points of the processing pipeline and a quality assessment on the final sample. 

The integrated plausibility check is an essential precondition for a successful completion of  an automated recognition scheme. It ensures that only samples which passed the check of a processing stage are handed over to the next stage. In the proposed pre-processing pipeline we implement three plausibility checks:
\begin{itemize}
	\item Segmentation: Analysis of the dominant components in the binary mask. Here the amount of dominant contours as well as their shape, size and position are analyzed. Also, the relative positions to each other are inspected. 
	\item Capturing: Evaluation of the fingerprint sharpness. A Sobel filter evaluates the sharpness of the processed gray scale fingerprint image. A square of 32$\times$32 pixels at the center of the image is considered. A histogram analysis then assesses the sharpness of the image. %
	\item Rotation, Cropping: Assessment of the fingerprint size. The size of the fingerprint image after the cropping stage shows if the fingerprint image is of sufficient quality. 
\end{itemize}
The combination of these plausibility checks has shown to be robust and accurate in our processing pipeline.
Every sample passing all three checks is considered as a candidate for the final sample. For every finger ID five samples are captured and processed. All five samples are finally assessed by NFIQ2.0 and the sample with the highest quality score is considered as the final sample. An assessment on the applicability of NFIQ2.0 on touchless fingerprint presented is shown in \cite{priesnitz2020touchless}. Especially for selecting the best sample from a series of the same subject, NFIQ2.0 is well-suited. 

\subsection{Feature Extraction and Comparison}
As mentioned earlier, the presented touchless fingerprint processing pipeline is designed in a way that obtained fingerprints are compatible with existing touch-based minutiae extractors and comparators. This enables the application of existing feature extraction and comparator modules within the proposed pipeline and facilitates a touch-based-to-touchless fingerprint comparison. Details of the employed feature extractor and comparator are provided in Section~\ref{sec:results}.

\section{Experimental Setup}
\label{sec:experimental_setup}
To benchmark our implemented app we conducted a data acquisition along with a usability study. Each volunteering subject first participated in a data acquisition session and then was asked to answer a questionnaire.

\subsection{Database Acquisition}
\label{sec:database_acquisition}
For the capturing of touchless samples two different setups were used: Firstly, a box-setup  simulates a predictable dark environment. Nevertheless, the subject was still able to place their fingers freely. Secondly, a tripod-setup simulates a fully free capturing setup where the instructor or the subject holds the capturing device \footnote{
The capture subjects handled the capturing devices without interaction of the instructor. This simulates an unattended capturing process and fulfilled the hygienic regulations during the database acquisition.
}.  

\begin{figure}[!t]
	\centering
		\subfigure[Age distribution]{\includegraphics[width=0.5\linewidth]{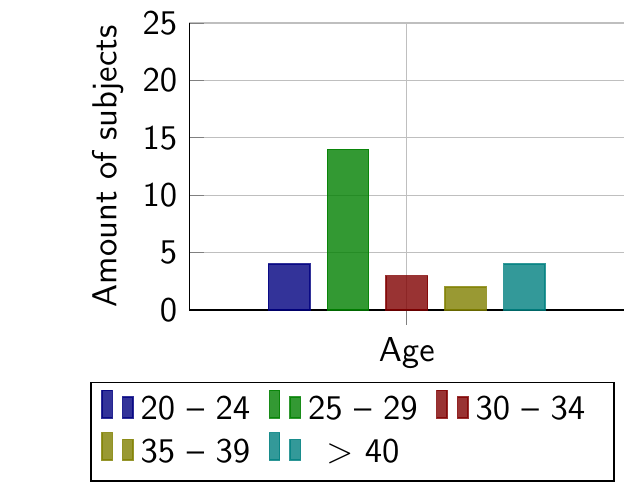}}\hfil
	\subfigure[Skin color distribution]{\includegraphics[width=0.5\linewidth]{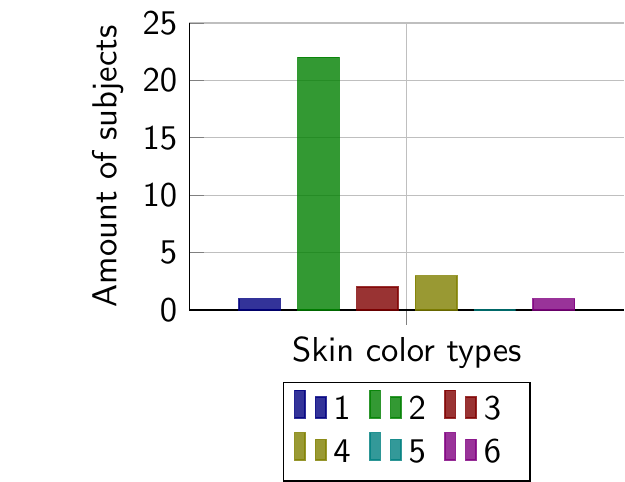}}\\
	\caption{Distribution of age and skin color according to Fitzpatrick metric \cite{fitzpatrick1988concept} of the subjects.}
	\label{fig:age_skin_color}
\end{figure}
For the touchless database capturing we used two different Smartphones: the Huawei P20 Pro (tripod setup) and the Google Pixel 4 (box setup). The finger images are captured with the highest possible resolution. The flashlight was turned on during the capturing process by default. 

Also, touch-based samples were captured to compare the results of the proposed setup against an established system. 
On every capturing device the four inner hand fingers (finger IDs 2 -- 5 and 7 -- 10 according to ISO/IEC 19794-4\cite{ISO19794-4}) were captured.  
The capturing of with three capturing devices shown in Figure~\ref{fig:capturing devices} was conducted in two rounds. 

To measure the biometric performance of the proposed system, we captured a database of 29 subjects. The age and skin color distribution can be seen in Figure~\ref{fig:age_skin_color}. Table~\ref{tab:setup} summarizes the database capturing method. 
During the capturing of one subject, Failure-To-Acquire (FTA) errors according to ISO/IEC 19795-1 \cite{ISO19795-1} occurred on both touchless capturing devices. Interestingly, this was most likely caused by the length of the subject's fingernails. For more information the reader is referred to Section~\ref{sec:fingernails}. In total, we captured and processed 1,360 fingerprints. 

\begin{table}[!t]
	\centering
	\footnotesize
	\caption{Overview on the NFIQ2.0 quality scores and the EER of all captured fingers (finger IDs 2 -- 5 and 7 -- 10) separated by sensors.}
	\label{tab:results_all}
	\begin{tabular}{|c|c|c|c|}
		\hline
		\textbf{Capturing device} & \textbf{Subset} & \begin{tabular}{@{}c@{}}\textbf{Avg. NFIQ2.0}\\ \textbf{score}\end{tabular} & \textbf{EER (\%)} \\\hline
		Touchless Box & all fingers & 44.80 ($\pm$ 13.51)  & 10.71 \\\hline
		Touchless Tripod & all fingers & 36.15 ($\pm$ 14.45)  & 30.41  \\\hline
		Touch-based & all fingers & 38.15 ($\pm$ 19.33
		)  & 8.19 \\\hline
	\end{tabular}
\end{table}
\begin{figure}[!t]
	\centering
	\subfigure[DET curves]{	\includegraphics[width=0.475\linewidth]{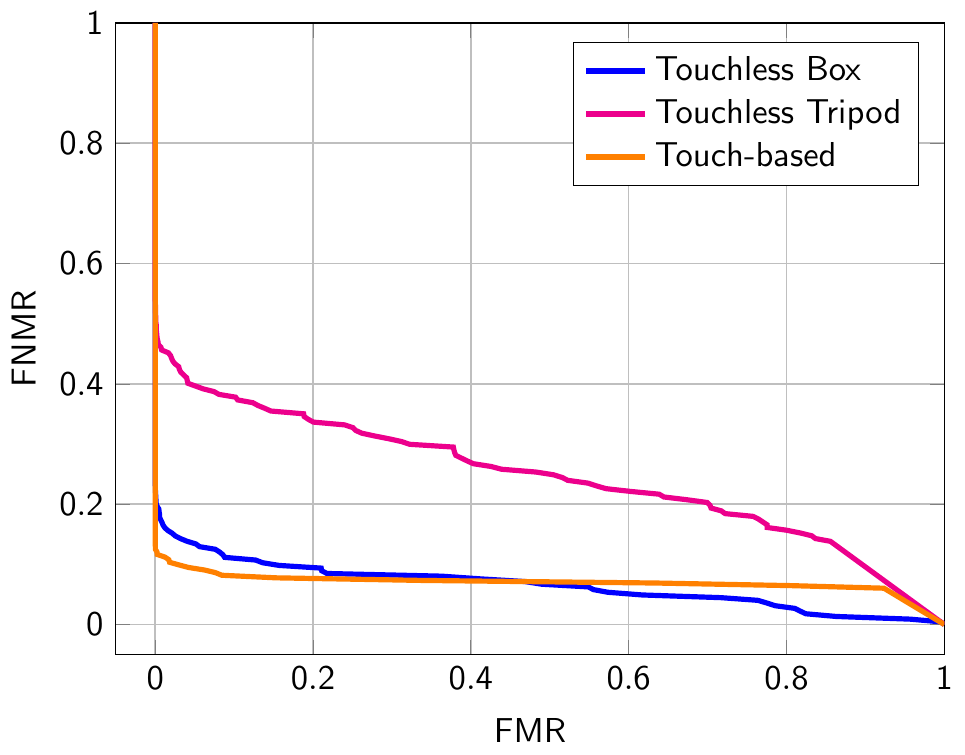}}\hfil
	\subfigure[Probability density functions of NFIQ2.0 scores]{\includegraphics[width=0.475\linewidth]{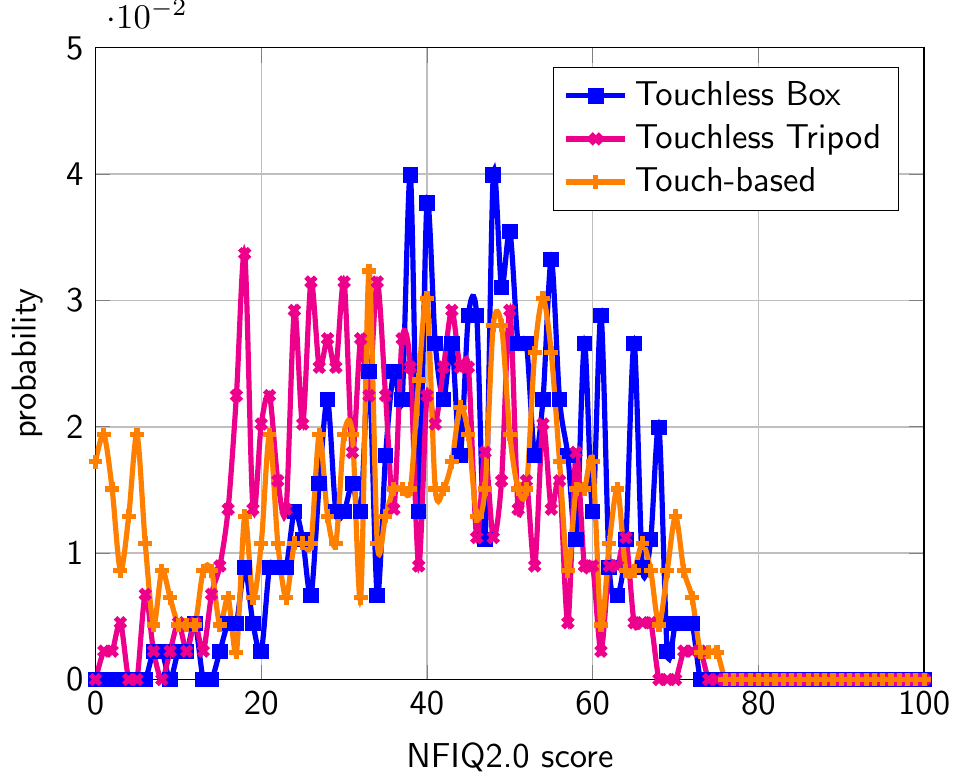}}
	\caption{NFIQ2.0 score distribution and biometric performance obtained from single finger comparisons.}\label{fig:score_dist}
\end{figure}

\subsection{Usability Study Design}\label{sec:study}
A usability study was conducted with each subject after they had interacted with the capturing devices. 
Each subject was asked about their individual preferences in terms of hygiene and convenience during the capturing process.
Parts of our usability study are based on \cite{furman_contactless_2017,weissenfeld2018contactless}.
We ensured that the questionnaire is as short and formulated as clearly as possible such that the participants understood all question correctly \cite{porst_fragebogen_2014}. %

The questionnaire is provided as supplemental material. It contains three parts: The first part contains questions about the subject's personal preferences. 
The Question 1.2b and 1.2c are aligned with Furman et al. \cite{furman_contactless_2017}. 
Here the different perceptions for personal hygiene before and during the COVID-19 pandemic were asked. The answer options of Question 1.5 were rated by the capture subjects using the Rohrmann scale \cite{rohrmann_empirische_1978} (strongly disagree, fairly disagree, undecided, fairly agree, strongly agree). 
The questions were intended to find out the subjects' perception regarding hygienic concerns during the fingerprint capturing process.

The second part of the questionnaire contains questions about the dedicated usability of a capturing device. The same questions were answered by the subject for both devices. 
This part was designed so that the same questions for both capturing devices were asked separately from each other in blocks. 
The intention behind this is to conduct comparisons between the different capturing devices.
Again, the Rohrman scale was used and sub-questions were arranged randomly.  %
In the last part of the questionnaire, the subjects were asked about the personal preference between both capturing devices. Here, the subjects had to choose one preferred capturing device.

\section{Results}
\label{sec:results}

\begin{figure*}[!t]
	\centering
	\subfigure[Touchless box setup]{\includegraphics[width=0.25\linewidth]{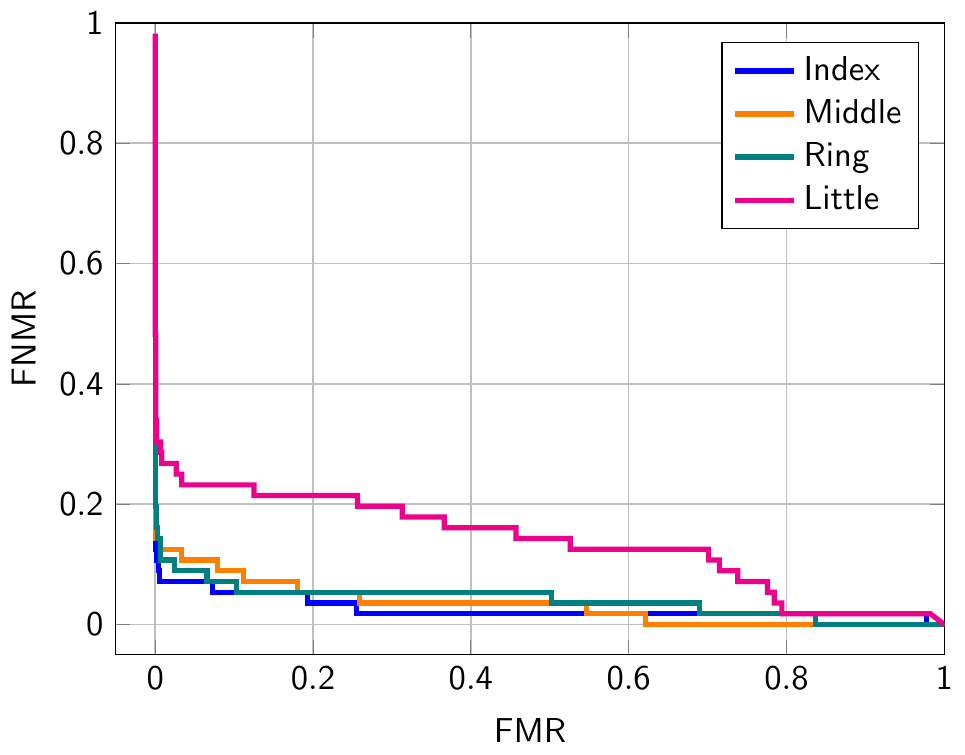}}\hfil
	\subfigure[Touchless tripod setup]{\includegraphics[width=0.25\linewidth]{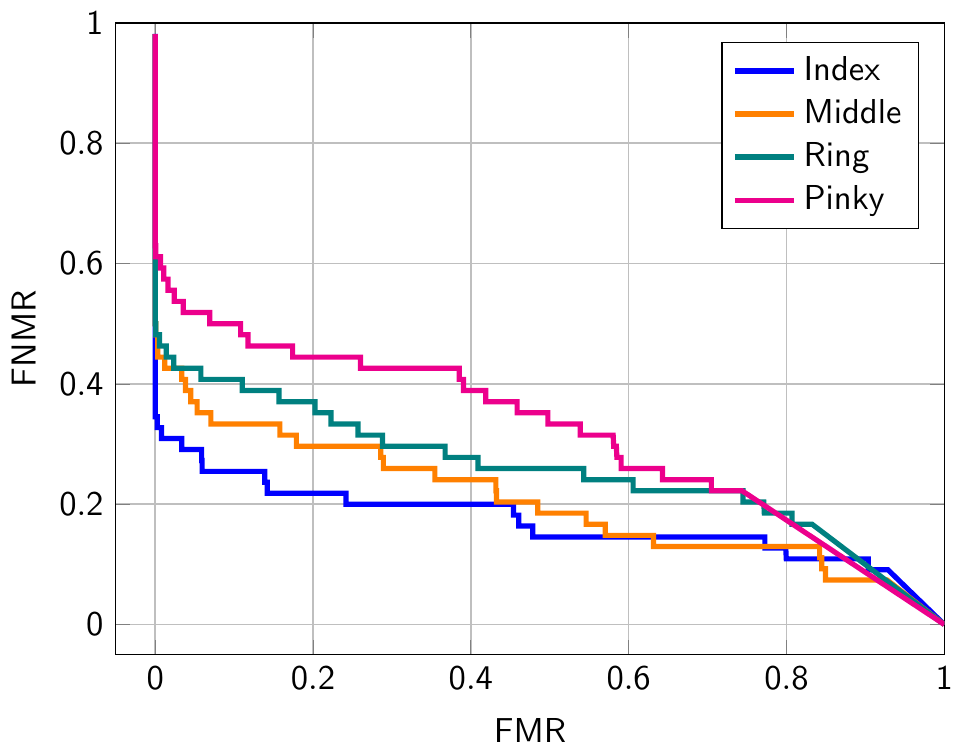}}\hfil
	\subfigure[Touch-based setup]{\includegraphics[width=0.25\linewidth]{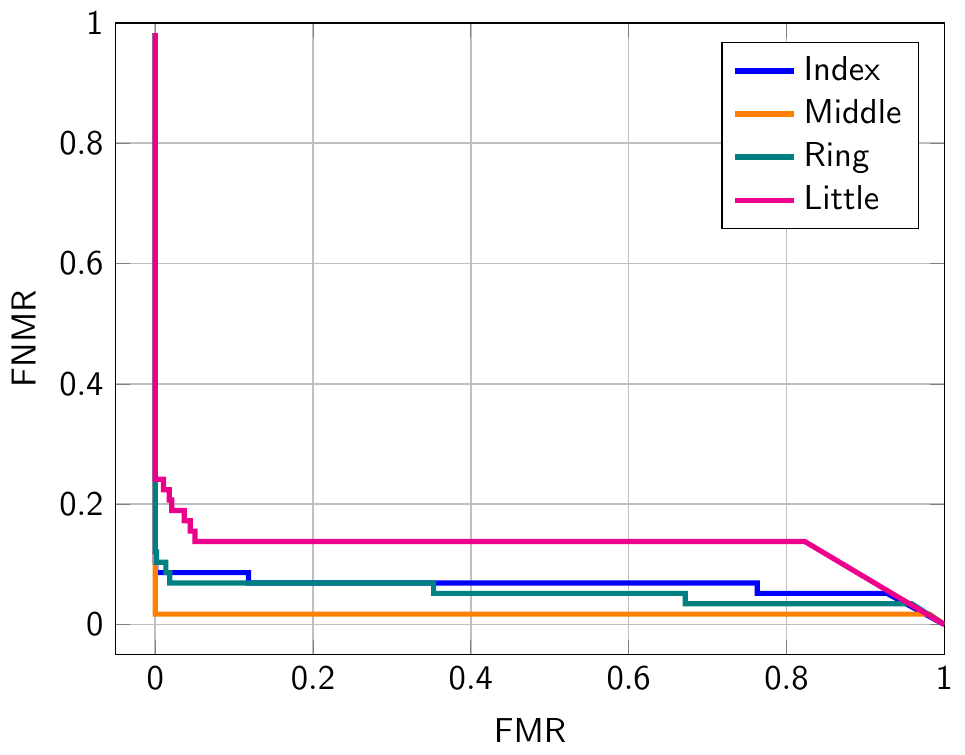}}\hfil
	\caption{DET curves obtained from individual finger comparisons: index fingers  (IDs 2, 7), middle fingers (IDs 3, 8), ring fingers (IDs 4, 9) and little fingers (IDs 5, 10).}\label{fig:det_individual_finger}
\end{figure*}

This section presents the biometric performance achieved by the entire recognition pipeline and the outcome of our usability study. 
\subsection{Biometric Performance}

\begin{figure}[!t]
	\centering
	\includegraphics[width=0.999999\linewidth]{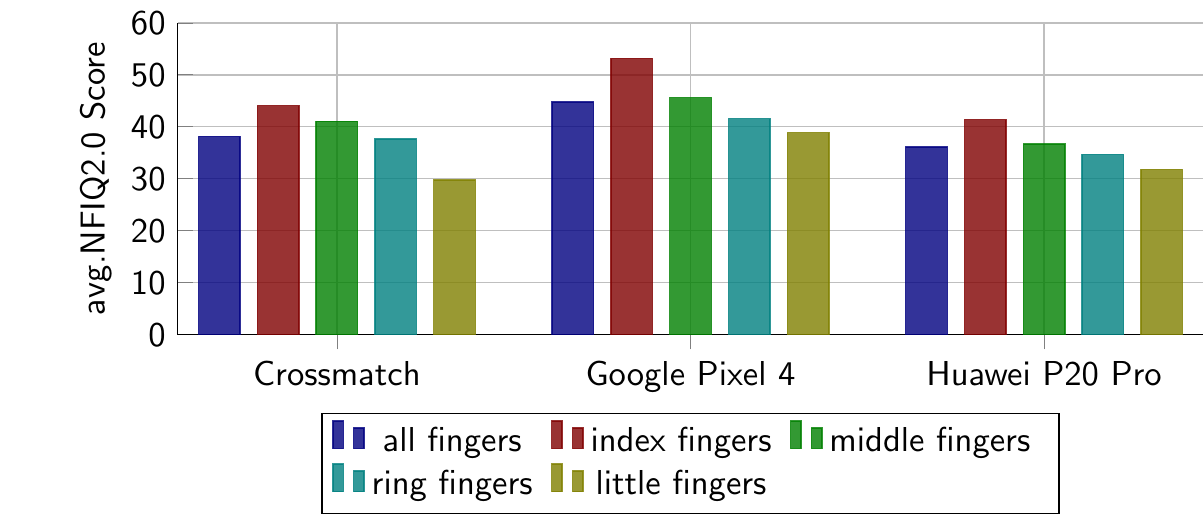}
	\caption{Averaged NFIQ2.0 scores obtained from the considered databases: average over all fingers (IDs 2 -- 4, 6 -- 10), index fingers  (IDs 2, 7), middle fingers (IDs 3, 8), ring fingers (IDs 4, 9) and little fingers (IDs 5, 10).}\label{fig:nfig_finger}
\end{figure}

\begin{table}[!t]
	\centering
	\footnotesize
	\caption{Overview on the NFIQ2.0 quality scores and the EER of individual fingers: index fingers  (IDs 2, 7), middle fingers (IDs 3, 8), ring fingers (IDs 4, 9) and little fingers (IDs 5, 10).}
	\label{tab:results_finger}
	\begin{tabular}{|c|c|c|c|}
		\hline
		\begin{tabular}{@{}c@{}} \textbf{Capturing}\\ \textbf{device} \end{tabular} & \textbf{Fingers} & \begin{tabular}{@{}c@{}} \textbf{Avg. NFIQ2.0}\\ \textbf{score} \end{tabular} & \begin{tabular}{@{}c@{}} \textbf{EER}\\ \textbf{(\%)} \end{tabular}\\\hline
		Touchless Box & index fingers & 53.16 ($\pm$ 11.27)  & 7.14 \\\hline
		Touchless Box & middle fingers & 45.59 ($\pm$ 11.06) & 8.91 \\\hline
		Touchless Box & ring fingers & 41.57 ($\pm$ 12.89) & 7.14 \\\hline
		Touchless Box & little fingers & 38.88 ($\pm$ 14.21) & 21.43 \\\hline
		Touchless Tripod  & index fingers & 41.38 ($\pm$ 14.29)  & 21.81  \\\hline
		Touchless Tripod & middle fingers &  36.68 ($\pm$ 13.01) & 28.58 \\\hline
		Touchless Tripod & ring fingers & 34.68 ($\pm$ 14.28) & 29.62 \\\hline
		Touchless Tripod & little fingers & 31.79 ($\pm$ 14.63) & 38.98 \\\hline
		Touch-based & index fingers & 44.06 ($\pm$ 17.53
		)  & 8.62  \\\hline
		Touch-based & middle fingers &  41.08 ($\pm$ 19.71
		) & 1.72 \\\hline
		Touch-based & ring fingers &  37.68 ($\pm$ 17.08
		) & 6.90 \\\hline
		Touch-based & little fingers & 29.78 ($\pm$ 19.94
		) & 13.79 \\\hline
	\end{tabular}
\end{table}

\begin{table}[!t]
	\centering
	\footnotesize
	\caption{Overview on the EER in a fingerprint fusion approach: Fusion over the 4 inner-hand fingers of the left hand (IDs 2 -- 4) and right hand (IDs 7 -- 10) fusing and 
		fusion over 8 fingers of both inner hands (IDs: 2 -- 4, 7 -- 10).}
	\label{tab:results_fusion}
	\begin{tabular}{|c|c|c|c|}
		\hline
		\textbf{Capturing device} & \textbf{Fusion approach} & \textbf{EER (\%)} \\\hline
		Touchless Box & 4 fingers & 5.36 \\\hline
		Touchless Box & 8 fingers & 0.00 \\\hline
		Touchless Tripod  & 4 fingers & 21.42 \\\hline
		Touchless Tripod & 8 fingers & 14.29 \\\hline
		Touch-based & 4 finger & 2.22 \\\hline
		Touch-based & 8 finger & 0.00 \\\hline
	\end{tabular}
\end{table}

\begin{figure}[!t]
	\centering
	\subfigure[Fusion over 4 fingers]{\label{subfig:det_fusion_4} \includegraphics[width=0.475\linewidth]{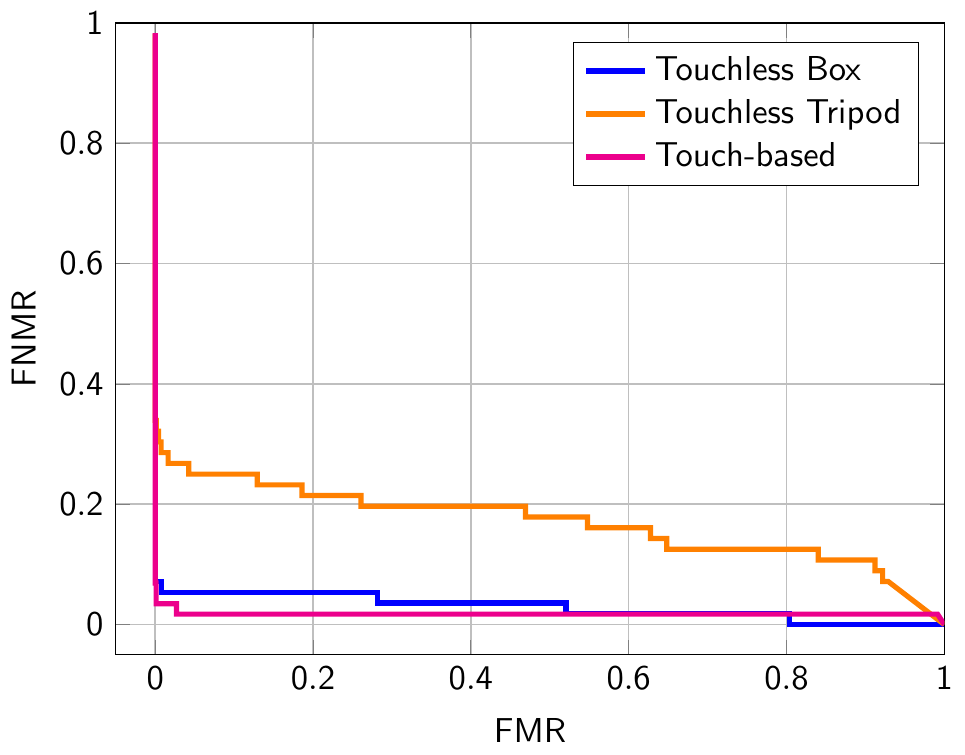}}\hfil
	\subfigure[Fusion over 8 fingers]{\label{subfig:det_fusion_8} \includegraphics[width=0.475\linewidth]{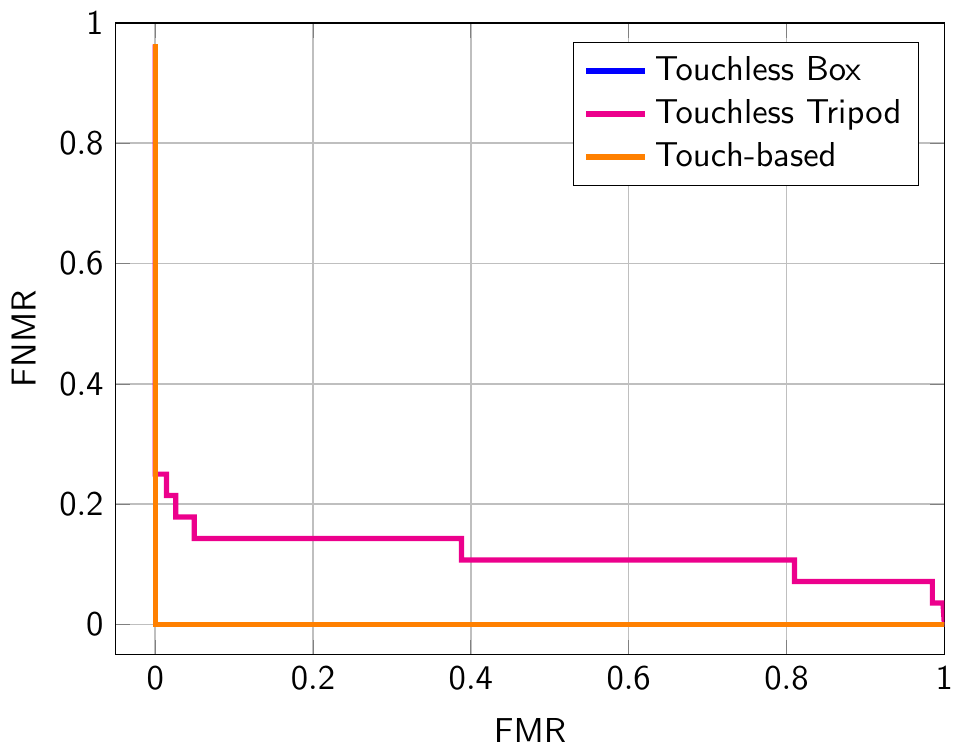}}\\
	\caption{DET curves obtained in a fingerprint fusion approach: Fusion over the 4 inner-hand fingers of the left hand (IDs 2 -- 4) and right hand (IDs 7 -- 10) fusing (a) and fusion over 8 fingers of both inner hands (IDs: 2 -- 4, 7 -- 10) (b).}
	\label{fig:fusion}
\end{figure}

In our experiments, we first estimate the distributions of NFIQ2.0 scores for the captured data set. Additionally, the biometric performance is evaluated employing open-source fingerprint recognition systems. The features (minutiae triplets -- 2-D location and angle) are extracted using a neural network-based approach. In particular, the feature extraction method of Tang \etal \cite{Tang-FingerNet-2017} is employed. For this feature extractor pre-trained models are made available by the authors. To compare extracted templates, a minutiae pairing and scoring algorithm of the sourceAFIS system of Va\v{z}an \cite{Vazan-SourceAFIS-2019} is used\footnote{The original algorithm uses minutiae quadruplets, \ie additionally considers the minutiae type (\eg ridge ending or bifurcation). Since only minutiae triplets are extracted by the used minutiae extractors, the algorithm has been modified to ignore the type information.}. 

In a first experiment, we compare the biometric performance on all fingers between the different sub data sets. From the Table~\ref{tab:results_all} we can see that the touchless box setup obtains an Equal Error Rate (EER) of $10.71\%$ which is comparable to the touch-based setup ($8.19\%$). Figure~\ref{fig:score_dist} (a) presents the corresponding Detection Error Trade-Off (DET) curve whereas Figure~\ref{fig:score_dist} (b) shows the probability density functions of NFIQ2.0 scores. 
In contrast, the performance of the open setup massively drops to an EER of $30.41\%$. The corresponding NFIQ2.0 scores do not reflect this drop  in terms of EER. Here, all three data sets have a comparable average score. 

In a second experiment, we compute the biometric performance for every finger separately\footnote{In this experiment we consider only the same finger IDs from a different subject as false~match.}. From Table~\ref{tab:results_finger} and Figure~\ref{fig:det_individual_finger} we can see that on all subsets the performance of the little finger drops compared to the other fingers. On the touch-based sub data set the middle finger has a much lower EER ($1.72\%$) than the rest. This could be because it might be easiest for users to apply the correct pressure to the middle finger. From Figure \ref{fig:nfig_finger} it is observable that there is only a small drop of NFIQ2.0 quality scores on the little finger. 

\begin{table}[!t]
	\footnotesize
	\centering
	\caption{Overview on the interoperability of different subset of the collected data: Comparison of fingerprints captured with different setups. All captured fingers (finger IDs 2 -- 5 and 7 -- 10) are considered. }\label{tab:interoperability}
	\begin{tabular}{|c|c|c|}
		\hline
		\textbf{Device A} & \textbf{Device B} & \textbf{EER (\%)} \\\hline
		Touchless Box & Touchless Tripod & 27.27 \\\hline
		Touchless Box & Touch-based & 15.71 \\\hline
		Touchless Tripod  & Touch-based & 32.02\\\hline
	\end{tabular}
\end{table}

\begin{figure}[!t]
	
	\centering
	\includegraphics[width=0.5\linewidth]{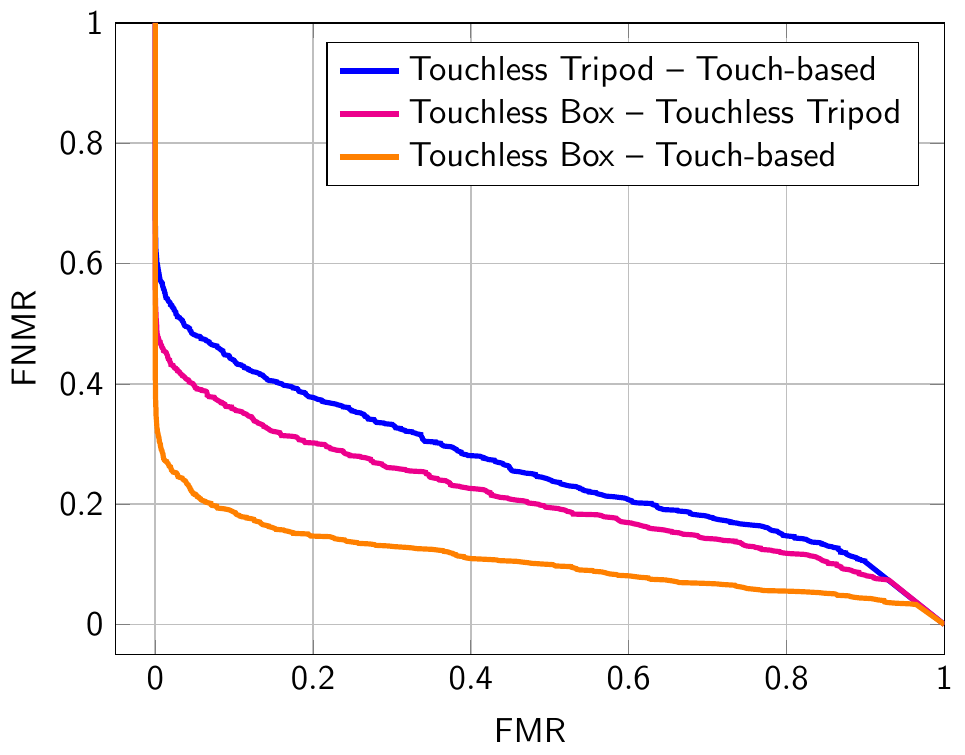}
	\caption{DET curves obtained from the interoperability of different subset of the collected data: Comparison of fingerprints captured with different setups. All captured fingers (finger IDs 2 -- 5 and 7 -- 10) are considered. }\label{fig:interoperability}
	
\end{figure}

\begin{figure*}[!t]
	\centering
	\includegraphics[width=0.68\textwidth]{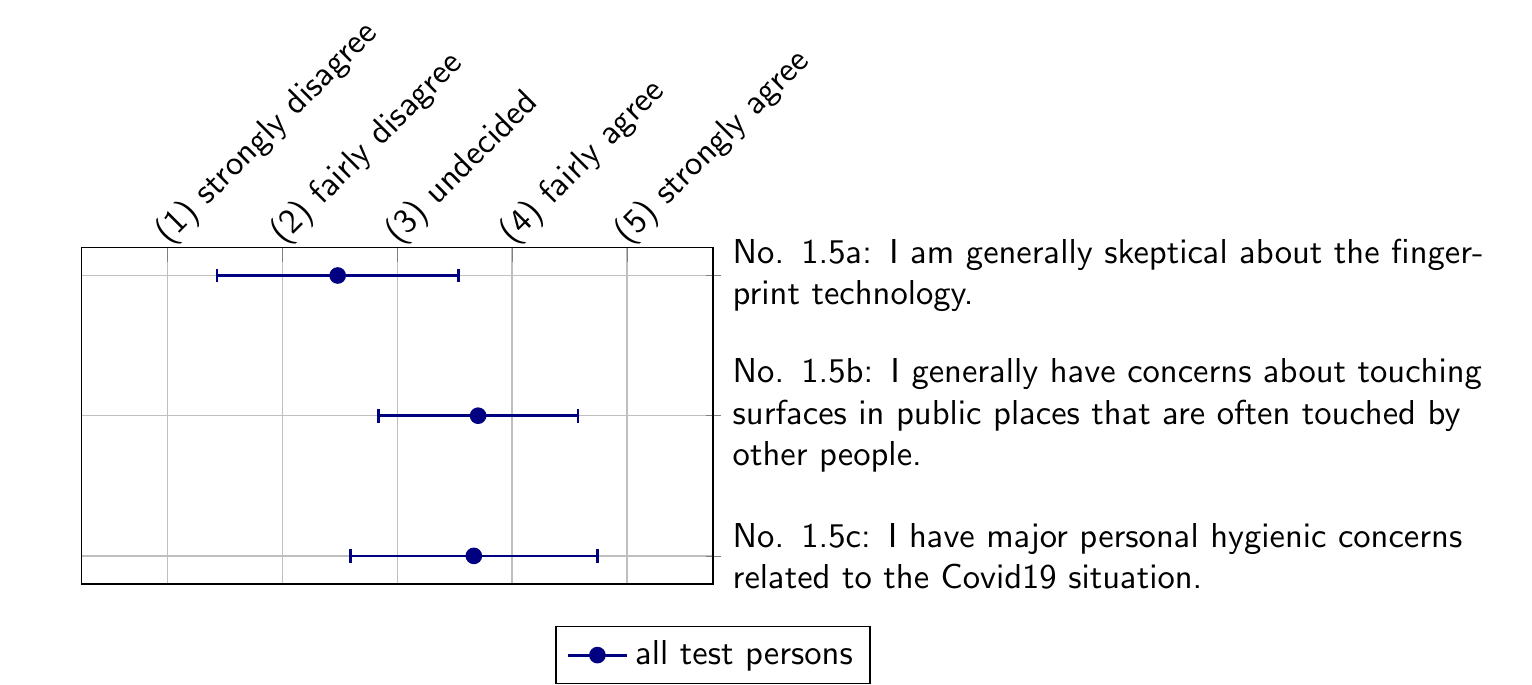}
	\caption{General assessment on fingerprint technology and hygienic concerns.}
	\label{fig:hygienicConcerns}
\end{figure*}
\begin{figure*}[!t]
	\centering
	\includegraphics[width=0.68\textwidth]{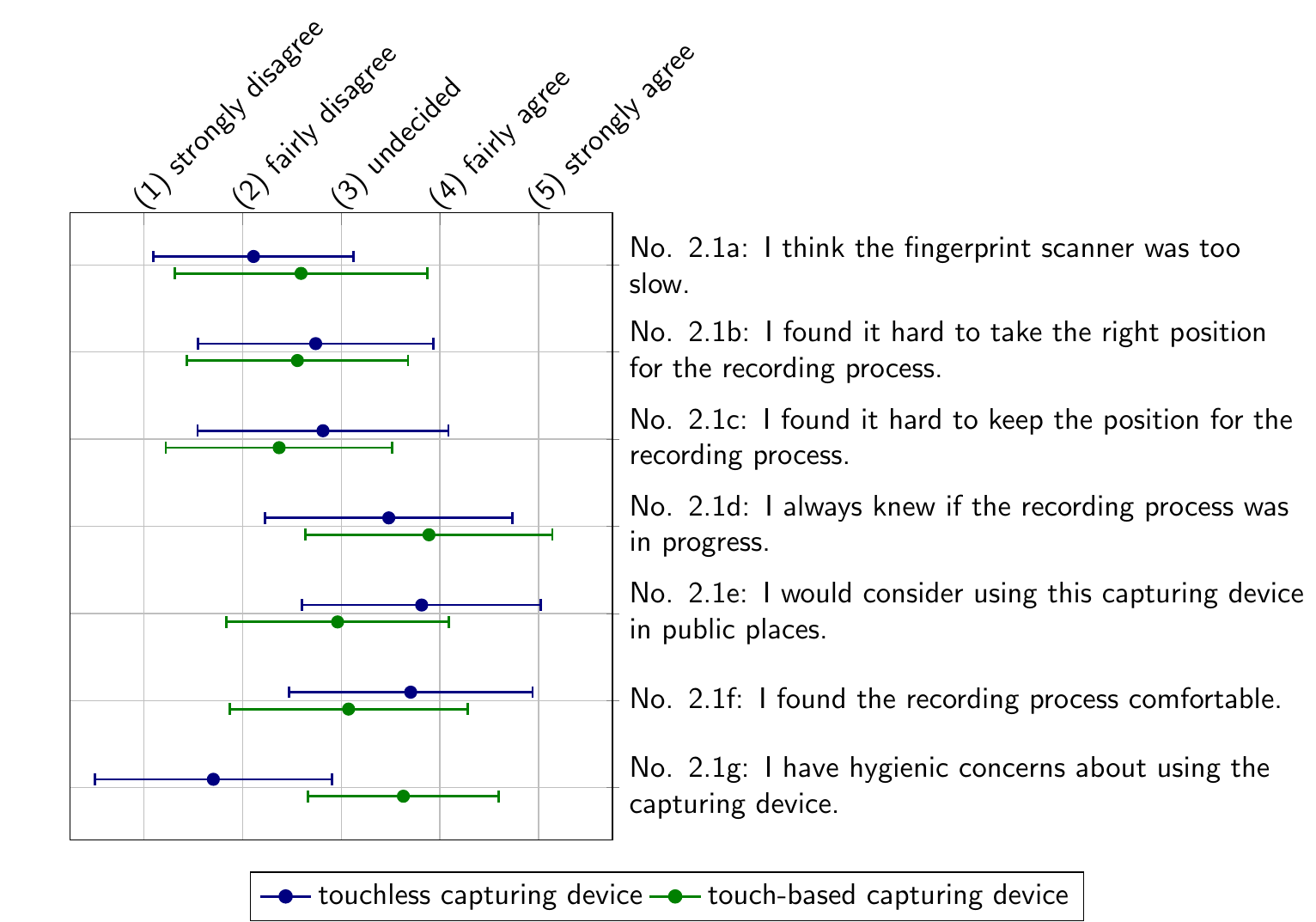}
	\caption{Usability assessment of the touchless and touch-based capturing device in comparison to each other.}
	\label{fig:vsRohrmann}
\end{figure*}

Further we applied a score level fusion on 4 and 8 fingers.
Obtained EERs are summarized in Table~\ref{tab:results_fusion}. As expected, the fusion improves the EER on all sub data sets. In particular, the fusion of 8 fingers shows a huge performance gain (see Figure~\ref{fig:fusion}). The box setup and the touch-based sensor show an EER of 0\% which means that matches and non-matches are completely separated. The open setup also achieves a considerably high performance gain trough the fusion. Here, the inclusion of all fingers makes the process much more robust especially in challenging environmental situations. 

In our last experiment we analyze the interoperability between the different subsets of the collected data. Table~\ref{tab:interoperability} summarizes the EERs achieved by comparing the samples of different setups and Figure~\ref{fig:interoperability} presents the corresponding DET curves. The touchless box setup shows a good interoperability to the touch-based setup ($15.71\%$). The EER of the open setup again significantly drops  ($27.27\%$ to touchless box and $32.02\%$ to touch-based). %

It should be noted, that all data sets, touchless as well as touch-based show a biometric performance which is inferior compared to the state-of-the-art. This is caused mainly by two reasons: On the one hand, all data sets were captured under realistic scenarios. The amount of instructions given to the capturing subject were limited to a minimum. Further, no re-capturing of poor quality samples took place. 
On the other hand, hygienic measures have an impact on biometric performance which is discussed in detail in Section~\ref{sec:covid_impact}.

\begin{figure*}[!t]
	\centering
	\includegraphics[width=0.625\linewidth]{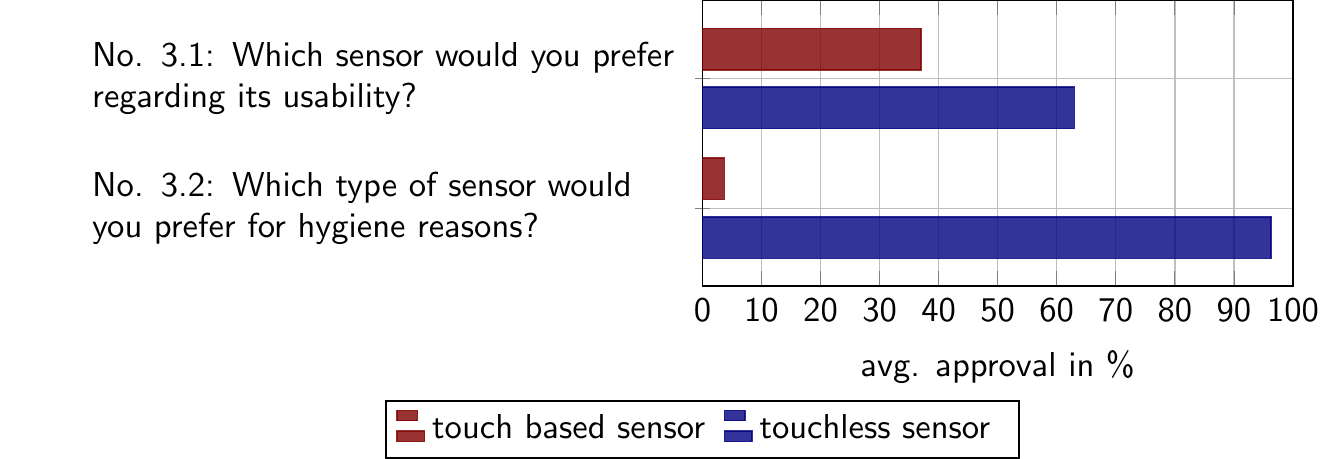}
	\caption{Comparative assessment of the capturing device type preference.}
	\label{fig:touchbasedvsless}
\end{figure*}

\subsection{Usability Study}\label{sec:studyResults}
We present the results of our usability study based on the questionnaire introduced in Section~\ref{sec:study}.
The questionnaire was answered by $27$ subjects ($8$ female, $19$ male). The subjects were between $22$ and $60$ years old (average age: $31.22$ years, median age: $28$ years). The age distribution is presented in Figure~\ref{fig:age_skin_color}. The exact result in terms of median and standard deviation is provided as supplemental material.
The majority of subjects have used professional fingerprint scanners before this study. A large proportion of the data subjects also use any type of fingerprint capturing device regularly at least once per week, e.g. to unlock mobile devices.

\begin{figure*}[!t]
	\centering
	\subfigure[Before COVID-19]{\label{subfig:XXX} \includegraphics[trim=-0.75cm 0 -0.75cm 0, clip,width=0.23\linewidth]{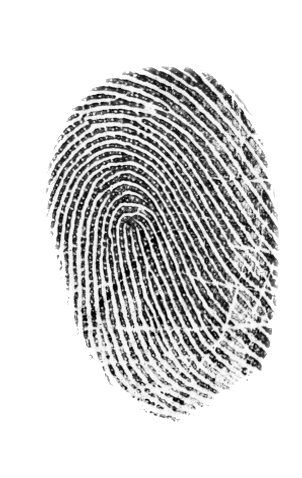}}\hfil
	\subfigure[During COVID-19]{\label{subfig:YYY} \includegraphics[trim=-0.75cm 0 -0.75cm 0, clip,width=0.23\linewidth]{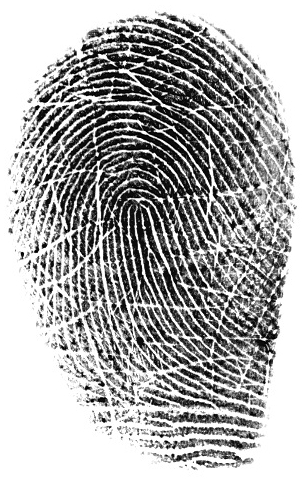}}\hfil
	\subfigure[Finger image]{\label{subfig:ZZZ} \includegraphics[trim=-0cm 1.9cm -0cm 0, clip,width=0.23\linewidth]{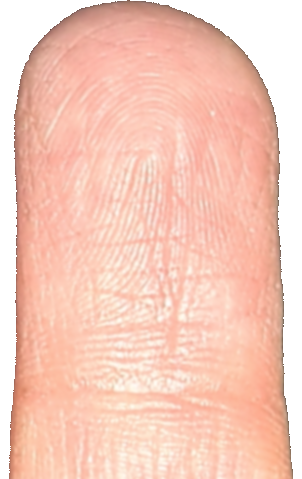}}\hfil
	\subfigure[Touchless sample]{\label{subfig:AAA} \includegraphics[trim=-0cm 1.9cm -0cm 0, clip,width=0.23\linewidth]{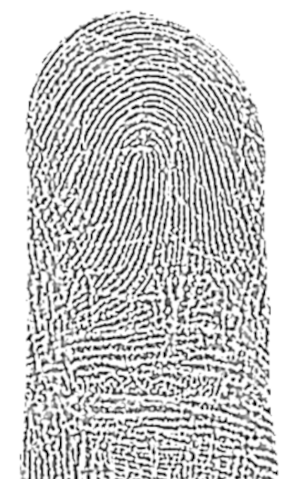}}
	\caption{Four samples of the same subject: Sample (a) was captured before the COVID-19 pandemic using a touch-based capturing device whereas samples (b -- d) were captured during the COVID-19 pandemic. Sample (a) and (b) were captured with the same capturing device whereas (c) and (d) are captured and processed using our method.}
	\label{fig:pre_post_corona}
\end{figure*}
Figure~\ref{fig:hygienicConcerns} presents the perceptions of the subjects regarding general hygiene. The subjects tend to have general concerns about touching surfaces in public places (Statement 1.5b). Moreover, the majority has personal concerns related to the COVID-19 pandemic (Statement 1.5c). 
From the small difference in terms of perception before and during the COVID-19 pandemic it can be observed that the pandemic has only a small influence on the general hygienic awareness of the tested subjects.

The usability assessment of the touchless and touch-based capturing devices is presented in Figure~\ref{fig:vsRohrmann}.
In most statements, both capturing devices were rated fairly similar.
The touchless capturing device has a slight advantage in terms of capturing speed (Statement 2.1a). 
The touch-based capturing device tends to be rated better in taking and keeping the capturing position during the whole process (Statements 2.1b and 2.1c). 
Also, the subjects found it slightly easier to assess if the capturing process is running (Statement 2.1d). 
Moreover, it can be observed that the subjects highly prefer the comfort of the touchless device (Statement 2.1f). Most notably, the subjects have much less hygienic concerns using the touchless device in public places (Statements 2.1.e and 2.1g). 
In these cases, a U-Test \cite{mann_test_1947} shows a two-sided significance with a level of $\alpha = 5\% $.

Figure~\ref{fig:touchbasedvsless} illustrates the comparative results. 
In a direct comparison of the different capturing device types the advantage of hygiene outweighs the disadvantages of hand positioning.
The slight majority of subjects would prefer the touchless capturing device over a touch-based one in terms of general usability (Question 3.1).
Considering hygienic aspects the vast majority would choose the touchless capturing device over the touch-based one (Question 3.2). This highly correlates to the assessment of hygienic concerns of the statement 1.5c. 

\begin{table*}[!t]
	\centering
	\footnotesize
	\caption{Average NFIQ2.0 scores and biometric performance obtained from touchless and touch-based databases including the fingerprint subcorpus of the MCYT Database \cite{1263277}, the FVC2006 Database \cite{CAPPELLI20077}, the Hong Kong Polytechnic University Contactless 2D to Contact-based 2D Fingerprint Images Database Version 1.0 \cite{kumar_hong_2017}.}
	\label{tab:EER_MCYT_FVC_PolyU}
	\begin{tabular}{|L{0.135\linewidth}|L{0.235\linewidth}|L{0.255\linewidth}|L{0.225\linewidth}|}
		\hline
		\textbf{Database} & \textbf{Subset} & \textbf{Avg. NFIQ2.0 score} & \textbf{EER (\%)} \\\hline
		\multirow{2}{*}{MCYT} & dp & 37.58 ($\pm$15.17) & 0.48 \\\cline{2-4}
		& pb & 33.02 ($\pm$13.99) & 1.35 \\\hline
		FVC06 & DB2-A & 36.07 ($\pm$9.07)  & 0.15 \\\hline
		\multirow{2}{*}{PolyU} 	& Contactless Session 1 & 47.71 ($\pm$10.86)  & 3.91 \\\cline{2-4}
		& Contactless Session 2 & 47.08 ($\pm$13.21)  & 3.17 \\\hline
		\multirow{2}{*}{Our database} & Touch-based  & 38.15 ($\pm$ 19.33
		)  & 8.19 \\\cline{2-4}
		& Touchless Box & 44.80 ($\pm$ 13.51)  & 10.71 \\\hline
	\end{tabular}
\end{table*}

\section{Impact of the COVID-19 Pandemic on Fingerprint Recognition}
\label{sec:covid_impact}
The accuracy of some biometric characteristics may be negatively impacted by the COVID-19 pandemic. The pandemic and its related measures have no direct impact on the operation of fingerprint recognition. Nevertheless, there are important factors that may indirectly reduce the recognition performance and user acceptance of fingerprint recognition.  

\subsection{Impact of Hand Disinfection on Biometric Performance}
The biometric performance drops due to dry and worn out fingertips. 
Olsen et al. \cite{olsen_fingerprint_2015} showed that the level of moisture has a significant impact on the biometric performance of touch-based fingerprint recognition systems. The authors tested five capturing devices with normal, wet and dry fingers. Especially dry fingers have been shown to be challenging.
Also, medical studies have shown that a frequent hand disinfection causes dermatological problems \cite{oconnell_case_2020, tan_contact_2020}. The disinfection liquids dry out the skin and cause chaps in the epidermis and dermis. 

Thus, we can infer that the regular hand disinfection leads to two interconnected problems which reduce the recognition performance: 
Dry fingers show low contrast during the capturing due to insufficient moisture. In addition, disinfection liquids lead to chaps on the finger surface. 
Figure~\ref{fig:pre_post_corona} shows touch-based fingerprints captured before the COVID-19 pandemic (a) and during the COVID-19 pandemic (b). Both samples were captured from the same subject using the same capturing device. 
It is observable that the sample (b) exhibits more impairments in the ridge-line pattern compared to sample (a). Moreover, the finger image (c) clearly shows chaps in the finger surface which are likely caused by hygienic measures. The processed touchless sample (d) shows these impairments, too.

Table~\ref{tab:EER_MCYT_FVC_PolyU} summarizes the biometric performance achieved on different databases using the proposed recognition pipeline. We can see that the biometric performance on the fingerprint subcorpus of the MCYT bimodal database \cite{1263277} and the Fingerprint Verification Contest 2006 (FVC06) \cite{CAPPELLI20077} show a good performance. Moreover, the touchless subset of the Hong Kong Polytechnic University Contactless 2D to Contact-based 2D Fingerprint Images Database Version 1.0 (PolyU) \cite{kumar_hong_2017} shows an competitive performance of $~ 3.50\%$ EER. Compared to these baselines the performance achieved on our database is inferior which is most likely the impact of the unconstrained acquisition scenario as well as the use of hand disinfection measures. 

\begin{figure*}[!t]
	\centering
	\subfigure[Sharpness assessment input]{\includegraphics[width=0.15\linewidth]{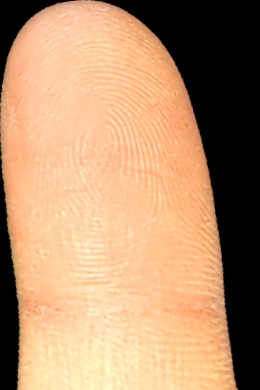}
		\includegraphics[width=0.15\linewidth]{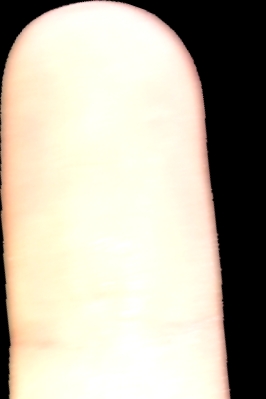}}\hfil
	\subfigure[Segmentation input]{\includegraphics[trim=0 147 0 147,clip,width=0.15\linewidth]{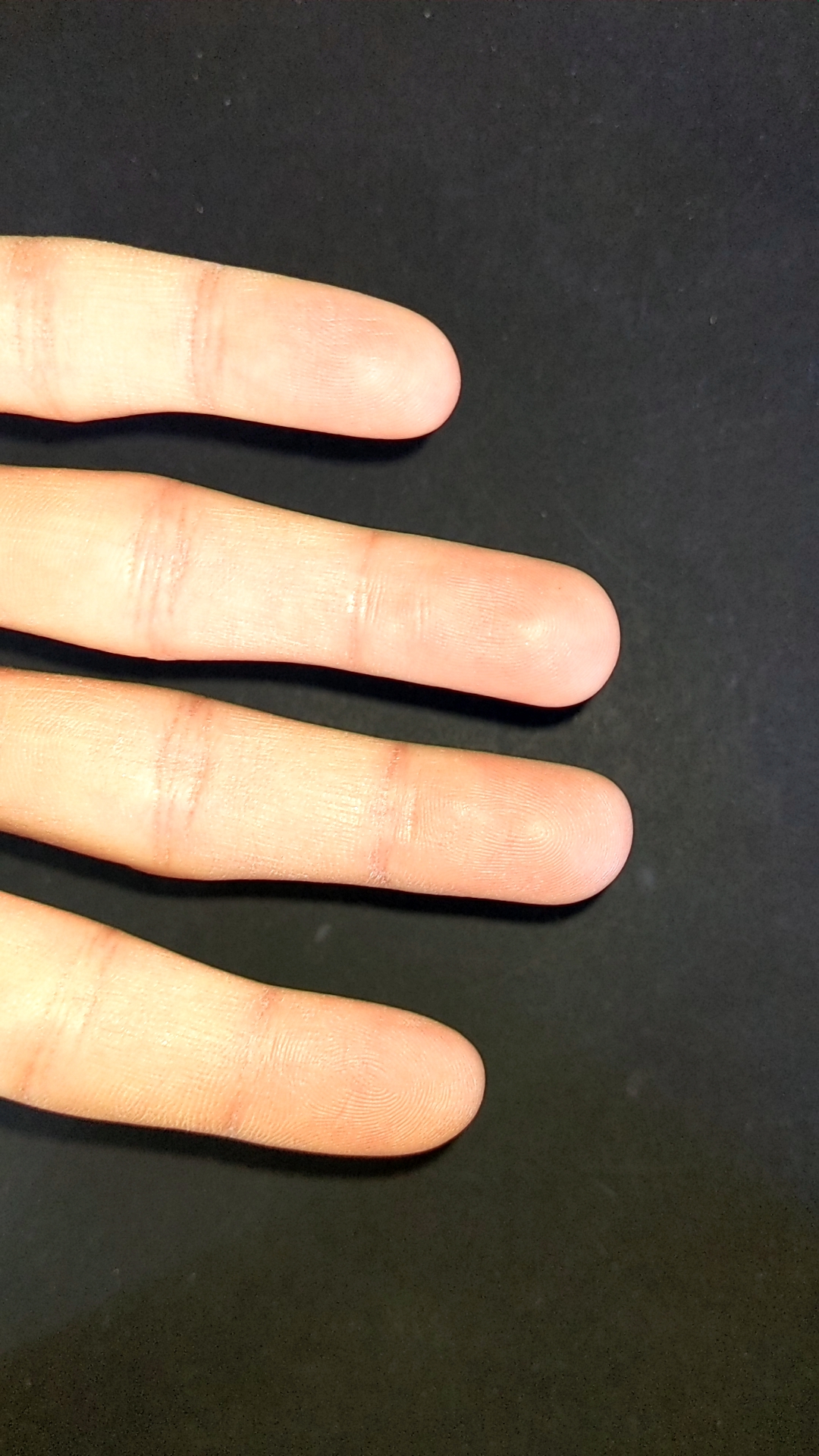}
		\includegraphics[trim=0 35 0 35,clip,width=0.15\linewidth]{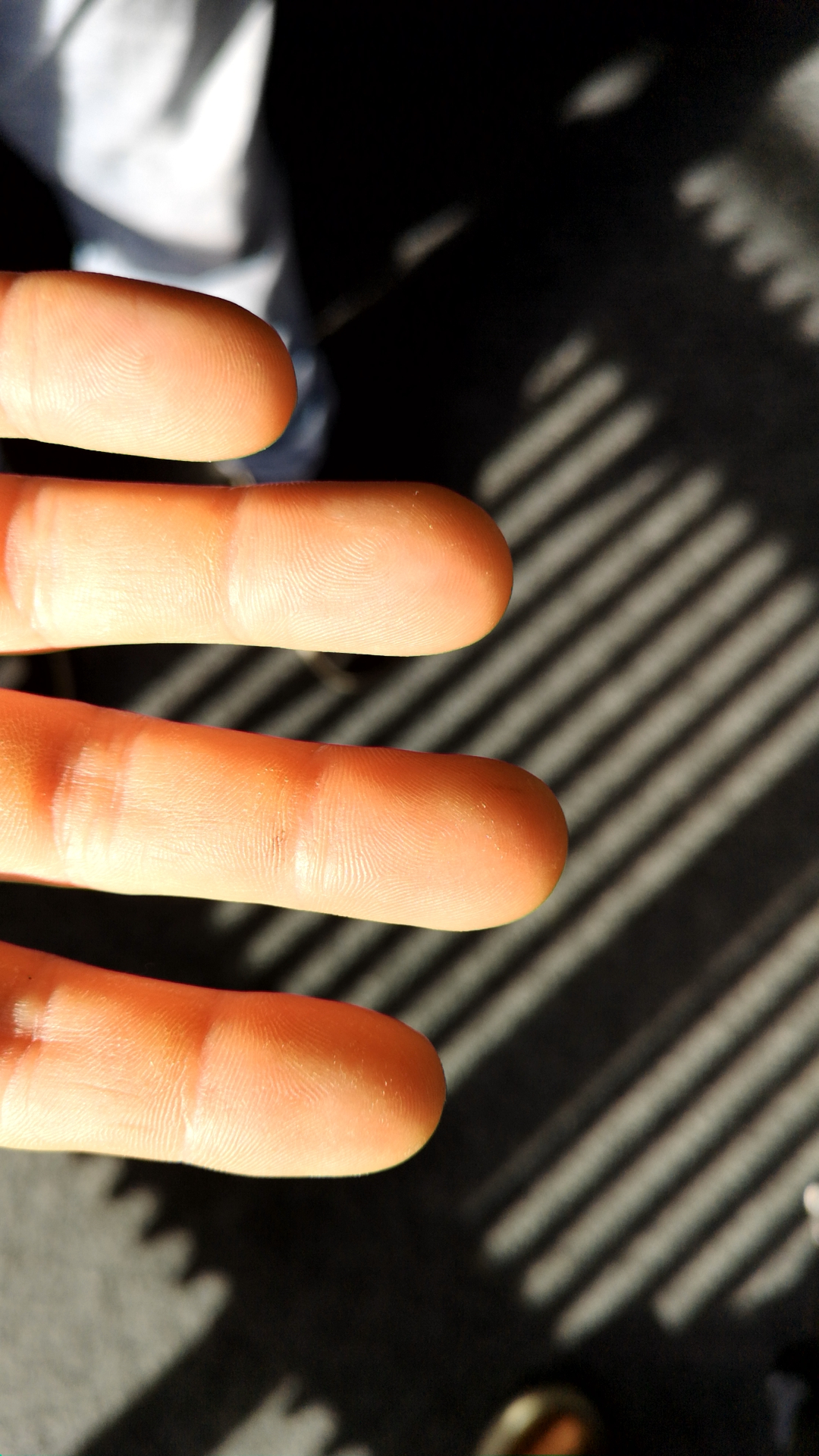}}\hfil
	\subfigure[CLAHE input]{\includegraphics[width=0.15\linewidth]{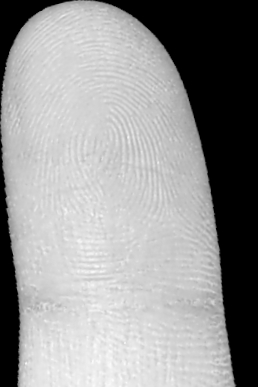}
		\includegraphics[width=0.15\linewidth]{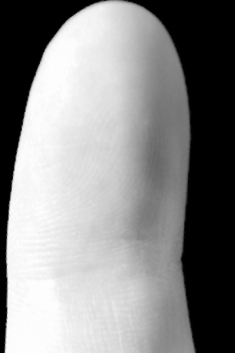}}\\
	\subfigure[Sharpness assessment result]{\includegraphics[width=0.15\linewidth]{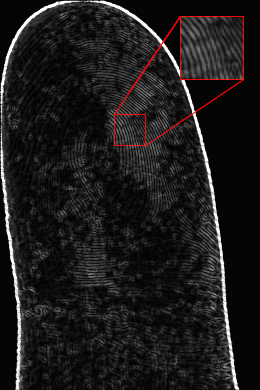}
		\includegraphics[width=0.15\linewidth]{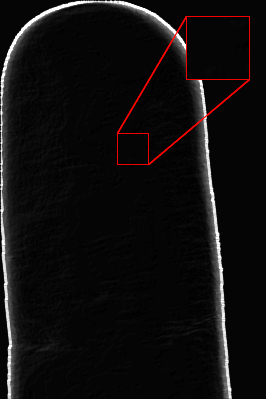}}\hfil
	\subfigure[Segmentation result]{\includegraphics[trim=0 35 0 35,clip,width=0.15\linewidth]{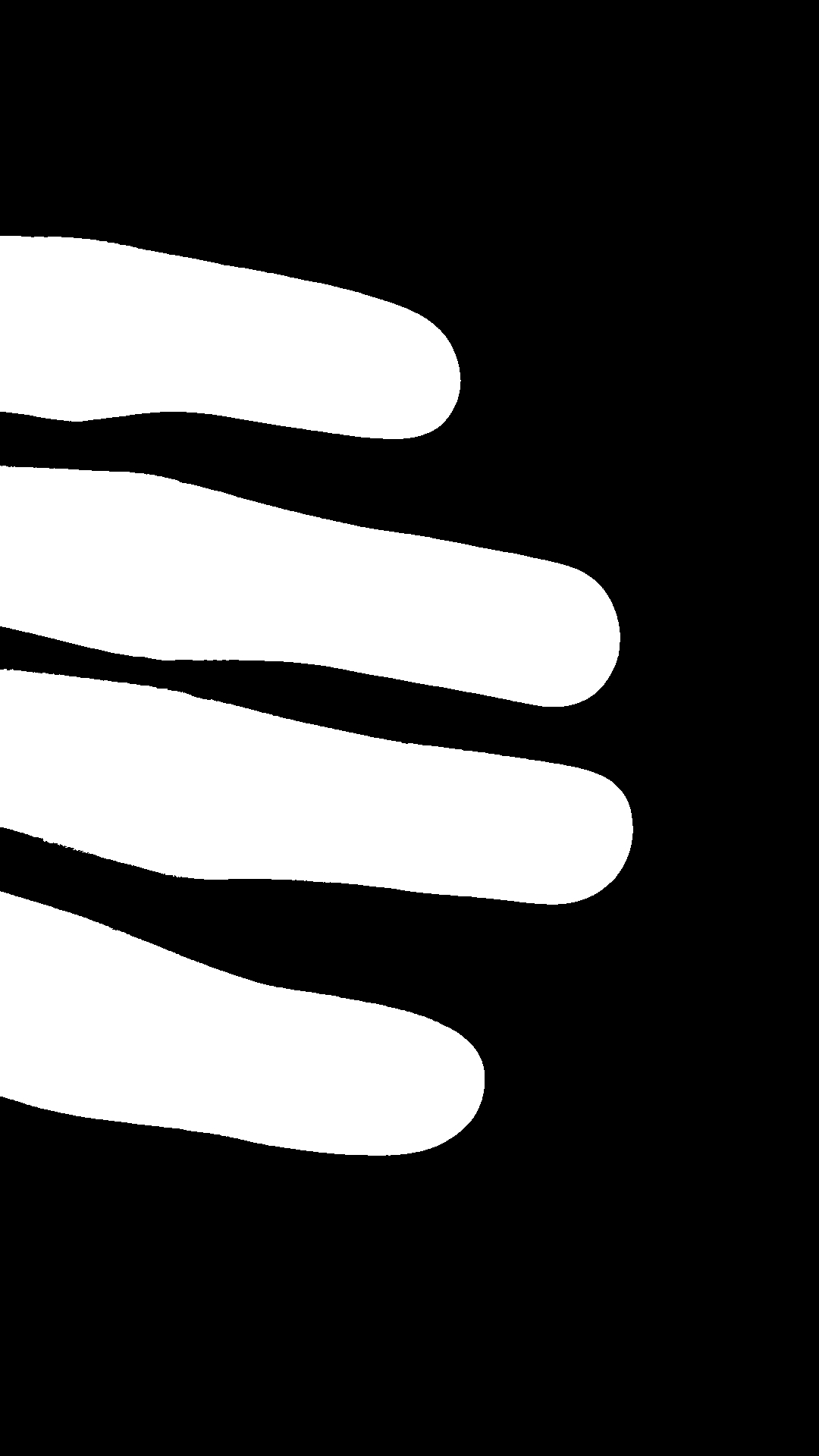}
		\includegraphics[trim=0 150 0 150,clip,width=0.15\linewidth]{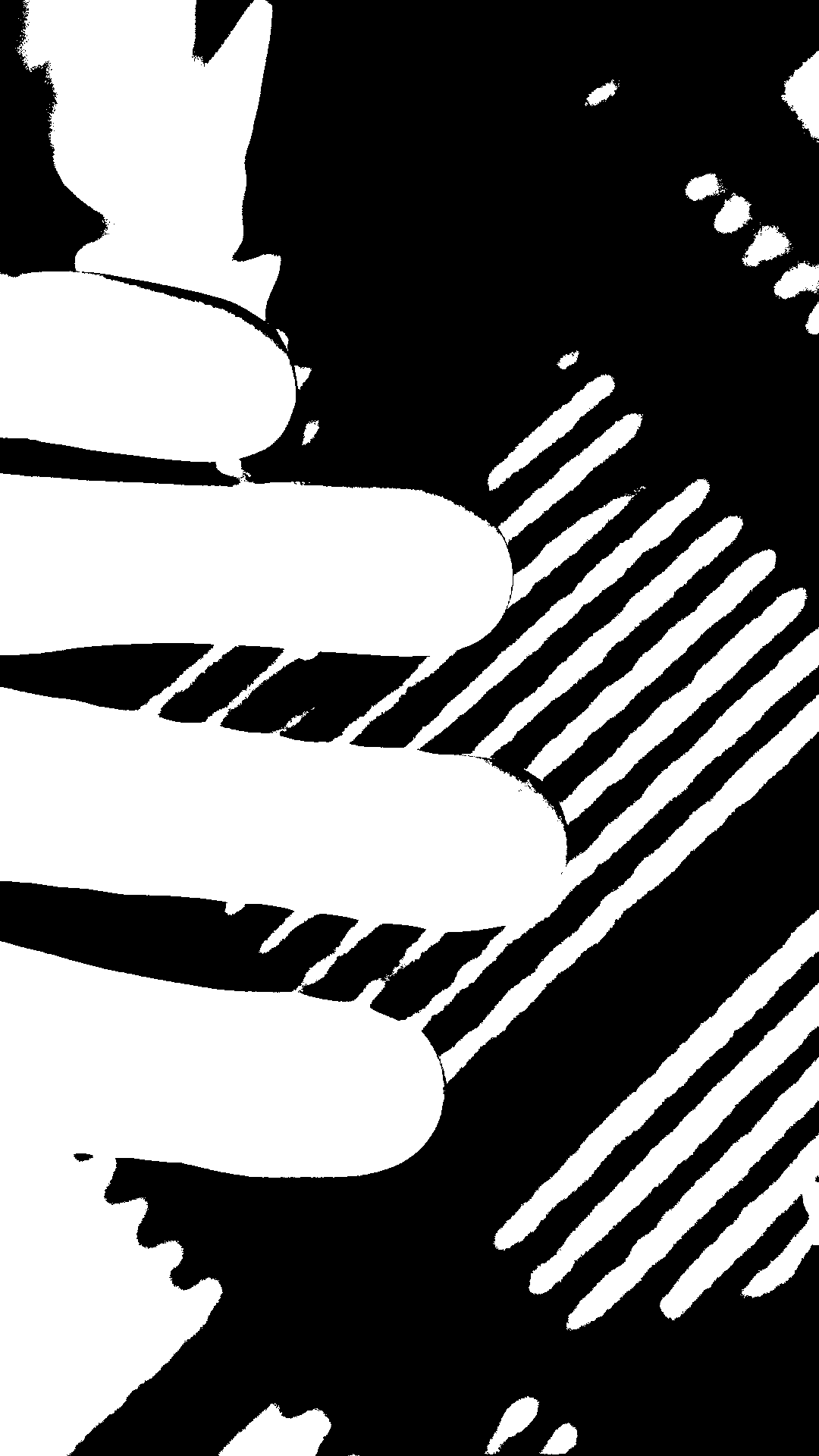}}\hfil
	\subfigure[CLAHE result]{\includegraphics[width=0.15\linewidth]{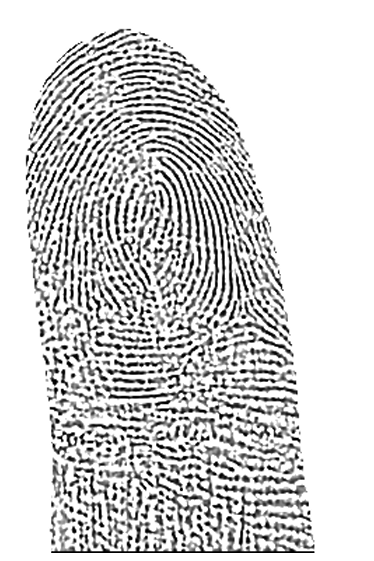}
		\includegraphics[width=0.15\linewidth]{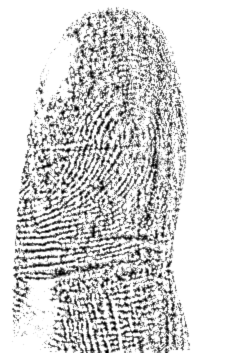}}\hfil
	\caption{Illustration of accurate and challenging input images and corresponding result images for  sharpness assessment (a, d), segmentation (b, e) and contrast adjustment (c, f). The left images of each block represent an accurate images the right one a challenging one.}\label{fig:sobel}
\end{figure*}
\subsection{User Acceptance} %
Viruses, e.g. SARS-CoV-2, have four main transmission routes: droplet, airborne, direct contact, and indirect contact via surfaces. In the last case an infected individual contaminates a surface by touching it. A susceptible individual who touches the surface afterwards has a high risk of infection via this indirect transmission route. Otter et al. \cite{otter_transmission_2016} present an overview on the transmission of different viruses (including SARS coronaviruses) via dry surfaces. The authors conclude that SARS coronaviruses can survive for extended periods on surfaces and for this reason form a high risk of infection. 

Especially in large scale implementations e.g. the Schengen Entry/Exit System (EES) \cite{union2019implementing}
were many individuals touch the surface of a capturing device, the users are exposed to a major risk of infection. 	
The only way of a safe touch-based fingerprint recognition in such application scenarios is to apply a disinfection of the capturing device after every subject. 

Nevertheless, the requirement of touching a surface can lower the user acceptance of touch-based fingerprint recognition. The results of our usability studies in Subsection~\ref{sec:studyResults} show that the asked individuals are fairly skeptical about touching capturing device surfaces in public places and that (in a direct comparison) they would prefer a touchless capturing device. For this reason the touchless capturing schemes could lead to a higher user acceptance. However, it should be noted that our tested group was very small and user acceptance is dependent to the capturing device design. 

\section{Implementation Aspects}
\label{sec:imeplmentation_aspects}
This section summarizes aspects which are considered beneficial for a practical implementation.

\subsection{Four-Finger Capturing}
As it has been shown in previous works, our proposed recognition pipeline demonstrates that it is possible to process four fingerprints from a continuous stream of input images. This requires a more elaborated processing but has two major advantages: 
\begin{itemize}
	\item Faster and more accurate recognition process: Due to a larger proportion of finger area in the image, focusing algorithms work more precisely. This results in less miss-focusing and segmentation issues.
	\item Improved biometric performance: The direct capturing of four fingerprints in one single capturing attempt is highly suitable for biometric fusion. As show in Table~\ref{tab:results_fusion} this lowers the EER without any additional capturing and very little additional processing. 
\end{itemize}

However, a major obstacle for touchless schemes is to capture the thumbs accurately and conveniently. In most environments, the best results are achieved with the inner-hand fingers facing upwards. This is ergonomically hard to achieve with thumbs.

\subsection{Automatic Capturing and On-Device Processing}
State-of-the-art smartphones feature powerful processing units which are capable to execute the described processing pipeline in a reasonable amount of time. We have shown that a robust and convenient capturing relies on an automatic capturing with integrated plausibility checks. Also, the amount of data which has to be transferred to a remote recognition workflow is reduced by an on-device processing and the recognition workflow can be based on standard components.

Especially in a biometric authentication scenario, it can be beneficial to integrate the feature extraction and comparison into the mobile device. In this case, an authentication of a previously enrolled subject can be implemented on a stand-alone device. 

\subsection{Environmental Influences}
Touchless fingerprint recognition in unconstrained environmental situations may be negatively affected by varying and heterogeneous influences.
In our experiments, we showed that our touchless setup performs rather good under a semi-controlled environment. The performance of the same recognition pipeline drastically drops in an uncontrolled environment. Here, it is observable that different stages of the processing pipeline suffer from challenging environments:
\begin{itemize}
	\item \emph{Focusing} of the hand area needs to be very accurate and fast in order to provide sharp finger images. Here, a focus point which is missed by a few millimeters causes a blurred and unusable image. Figure~\ref{fig:sobel}(a, d) illustrate the difference between a sharp finger image and a slightly de-focused image with help of a Sobel filter. Additionally, the focus has to follow the hand movement in order to achieve a continuous stream of sharp images. The focus of our tested devices tend to fail under challenging illuminations which was not the case in the constrained environment.
	\item \emph{Segmentation, rotation and finger separation} rely on a binary mask in which the hand area is clearly separated from the background. Figure~\ref{fig:sobel}(b, e) show examples of a successful and unsuccessful segmentation. Impurities in the segmentation mask lead to connected areas between the fingertips and artifacts at the border region of the image. This causes inaccurate detection and separation of the fingertips and incorrect rotation results. Because of heterogeneous background this in more often the case in unconstrained setups.
	\item \emph{Finger image enhancement} using the CLAHE algorithm normalizes dark and bright areas on the finger image. From Figure~\ref{fig:sobel}(c, e) we can see that this also works on samples of high contrast. Nevertheless, the results on challenging images may become more blurry. 
\end{itemize}
The discussed challenges lead to a longer capturing time and for this reason lower the usability and user acceptance. Further, the recognition performance in unconstrained environments is limited. Here, a weighing between usability and performance should be done based on the intended use case of the capturing device.
The quality assessments implemented in our scheme detect these circumstances and discard finger images with said shortcomings. More elaborated methods could directly adapt to challenging images, e.g. by changing the focusing method or segmentation scheme. This approach could lead to more robustness and hence improved usability in different environments.

\subsection{Feature Extraction Strategies}
\begin{figure}[t]
	\centering
	\includegraphics[width=0.99\linewidth]{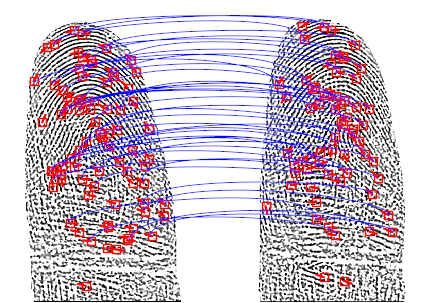}
	\caption{Illustration of a minutiae-based comparison of two touchless fingerprint samples. The features are extracted using the method described in Section \ref{sec:experimental_setup}. The blue lines indicate mated minutiae.}
	\label{fig:222mntcmp}
\end{figure}
Feature extraction techniques are vital to achieve a high biometric performance. In our experiments, we used an open-source feature extractor which is able to process touch-based and touchless samples. Figure~\ref{fig:222mntcmp} shows an example of this minutiae-based feature extraction and comparison scheme. With this method we were able to test the interoperability between capturing device types. Nevertheless, the overall performance may be improved by more sophisticated methods, e.g. commercial of-the-shelf-systems like the VeriFinger SDK \cite{verifinger_neuro_2010}. %

Touchless fingerprint images do not correspond to the standardized 500dpi resolution of touch-based capturing devices because of a varying distance between capturing device and finger. Here a feature extractor which is robust against these resolution differences or a dedicated processing step is beneficial. 

Also, dedicated touchless feature extraction methods can increase the performance as shown in \cite{sankaran2015smartphone, vyas_collaborative_2019}. Here the authors are able to tune their feature extractor to their capturing and processing and propose an end-to-end recognition system.

\subsection{Visual Instruction}
According to the presented results in Subsection~\ref{sec:studyResults}, the visual feedback of the touchless capturing device have also been rated to be inferior compared to the touch-based one. Here the smartphone display is well-suited to show further information about the capturing process. Additionally, an actionable feedback can be given on the positioning of the fingers, as suggested in \cite{carney2017multi-finger}.

\subsection{Robust Capturing of different Skin Colors and Finger Characteristics}
\label{sec:fingernails}
An important implementation aspect of biometric systems is that they must not discriminate certain user groups based on skin color or other characteristics. 
During our database capturing subjects of different skin color types were successfully captured. Nevertheless, it must be noted that the amount of subjects is too small to make a general statement about the fairness of the presented approach. 

As already mentioned in Section~\ref{sec:database_acquisition}, we observed one single failure to acquire (FTA) during our database acquisition. Most likely the cause for this was that the subject had very long fingernails which were segmented as finger area. Here the plausibility check during the segmentation failed and a capturing of the subject was not possible. 
To overcome this flaw a fingernail detection could be implemented in the segmentation workflow.

\section{Conclusion}
\label{sec:conclusion}
In this work, we proposed a fingerprint recognition workflow for state-of-the-art smartphones. The method is able to automatically capture the four inner-hand fingers of a subject and process them to separated fingerprint images. 
With this scheme we captured a database of 1,360 fingerprints from 29 subjects. Here we used two different setups: a box setup with  constrained environmental influences and a tripod setup.  Additionally, we captured touch-based fingerprints as baseline.
During a usability study after the capturing the subjects were asked about their experience with the different capturing device types. 

Our investigations show that the overall biometric performance of the touchless box setup is comparable to the touch-based baseline whereas the unconstrained touchless tripod setup shows inferior results. All setups benefit from a biometric fusion. A further experiment on the interoperability between touchless and touch-based samples (box setup) shows that the performance  drops only slightly. 

The presented usability study shows that the majority of users prefer a touchless recognition system over a touch-based one for hygienic reasons. Also, the usability of the touchless capturing device was seen as slightly better.  Nevertheless, the user experience of the tested touchless devices can be further improved. 

The COVID-19 pandemic also has an influence on  the performance and acceptance of fingerprint recognition systems. Here hygienic measures lower the recognition performance and users are more concerned touching surfaces in public areas. 

Our proposed method forms a baseline for a mobile automatic touchless fingerprint recognition system and is made publicly available. Researchers are encouraged to integrate their algorithms into our system and contribute to an more accurate, robust and secure touchless fingerprint recognition scheme.

\section*{Acknowledgments}
The authors acknowledge the financial support by the Federal Ministry of Education and Research of Germany in the framework of MEDIAN (FKZ 13N14798). This research work has been partially funded by the German Federal Ministry of Education and Research and the Hessian Ministry of Higher Education, Research, Science and the Arts within their joint support of the National Research Center for Applied Cybersecurity ATHENE.
\bibliography{mybibfile}

\begin{thebibliography}{10}
\expandafter\ifx\csname url\endcsname\relax
  \def\url#1{\texttt{#1}}\fi
\expandafter\ifx\csname urlprefix\endcsname\relax\def\urlprefix{URL }\fi
\expandafter\ifx\csname href\endcsname\relax
  \def\href#1#2{#2} \def\path#1{#1}\fi

\bibitem{okereafor_fingerprint_2020}
K.~Okereafor, I.~Ekong, I.~Okon~Markson, K.~Enwere,
  \href{http://biomedeng.jmir.org/2020/1/e19623/}{Fingerprint {Biometric}
  {System} {Hygiene} and the {Risk} of {COVID}-19 {Transmission}}, JMIR
  Biomedical Engineering 5~(1) (2020) e19623.
\newblock \href {https://doi.org/10.2196/19623} {\path{doi:10.2196/19623}}.
\newline\urlprefix\url{http://biomedeng.jmir.org/2020/1/e19623/}

\bibitem{hiew2007digital}
B.~Y. {Hiew}, A.~B.~J. {Teoh}, Y.~H. {Pang}, Digital camera based fingerprint
  recognition, in: International Conference on Telecommunications and Malaysia
  International Conference on Communications, 2007, pp. 676--681.

\bibitem{piuri2008fingerprint}
V.~Piuri, F.~Scotti, Fingerprint {Biometrics} via {Low}-cost {capturing
  devices} and {Webcams}, in: {Second} {International} {Conference} on
  {Biometrics}: {Theory}, {Applications} and {Systems} (BTAS), 2008, pp. 1--6.

\bibitem{wang2009novel}
L.~Wang, R.~H.~A. El-Maksoud, J.~M. Sasian, W.~P. Kuhn, K.~Gee, V.~S. Valencia,
  A novel contactless aliveness-testing (cat) fingerprint capturing device, in:
  Novel Optical Systems Design and Optimization XII, Vol. 7429, 2009, p.
  742915.

\bibitem{kumar2011contactless}
A.~{Kumar}, Y.~{Zhou}, Contactless fingerprint identification using level zero
  features, in: {Conference} on {Computer} {Vision} and {Pattern} {Recognition}
  Workshops (CVPRW), 2011, pp. 114--119.

\bibitem{derawi2011fingerprint}
M.~O. Derawi, B.~Yang, C.~Busch, Fingerprint recognition with embedded cameras
  on mobile phones, in: Security and Privacy in Mobile Information and
  Communication Systems (ICST), 2012, pp. 136--147.

\bibitem{noh2011touchless}
D.~Noh, H.~Choi, J.~Kim, Touchless capturing device capturing five fingerprint
  images by one rotating camera, Optical Engineering 50~(11) (2011) 113202.

\bibitem{stein2013video-based}
C.~Stein, V.~Bouatou, C.~Busch, Video-based fingerphoto recognition with
  anti-spoofing techniques with smartphone cameras, in: {International}
  {Conference} of the {Biometric} {Special} {Interest} {Group} ({BIOSIG}),
  2013, pp. 1--12.

\bibitem{raghavendra2014low-cost}
R.~Raghavendra, K.~B. Raja, J.~Surbiryala, C.~Busch, A low-cost multimodal
  biometric capturing device to capture finger vein and fingerprint, IEEE
  International Joint Conference on Biometrics (2014) 1--7.

\bibitem{tiwari2015touch-less}
K.~Tiwari, P.~Gupta, A touch-less fingerphoto recognition system for mobile
  hand-held devices, in: {International} {Conference} on {Biometrics} ({ICB}),
  2015, pp. 151--156.

\bibitem{sankaran2015smartphone}
A.~Sankaran, A.~Malhotra, A.~Mittal, M.~Vatsa, R.~Singh, On smartphone camera
  based fingerphoto authentication, in: 7th {International} {Conference} on
  {Biometrics} {Theory}, {Applications} and {Systems} ({BTAS}), 2015, pp. 1--7.

\bibitem{carney2017multi-finger}
L.~A. Carney, J.~Kane, J.~F. Mather, A.~Othman, A.~G. Simpson, A.~Tavanai,
  R.~A. Tyson, Y.~Xue, A multi-finger touchless fingerprinting system: Mobile
  fingerphoto and legacy database interoperability, in: 4th International
  Conference on Biomedical and Bioinformatics Engineering (ICBBE), 2017, p.
  139–147.

\bibitem{deb2018matching}
D.~Deb, T.~Chugh, J.~Engelsma, K.~Cao, N.~Nain, J.~Kendall, A.~K. Jain,
  Matching fingerphotos to slap fingerprint images, arXiv preprint
  arXiv:1804.08122 (2018).

\bibitem{weissenfeld2018contactless}
A.~Weissenfeld, B.~Strobl, F.~Daubner, Contactless finger and face capturing on
  a secure handheld embedded device, in: 2018 {Design}, {Automation} {Test} in
  {Europe} {Conference} {Exhibition} ({DATE}), 2018, pp. 1321--1326.

\bibitem{birajadar_towards_2019}
P.~Birajadar, M.~Haria, P.~Kulkarni, S.~Gupta, P.~Joshi, B.~Singh, V.~Gadre,
  Towards smartphone-based touchless fingerprint recognition, Sādhanā 44~(7)
  (2019) 161.

\bibitem{chen20063d}
Y.~Chen, G.~Parziale, E.~Diaz-Santana, A.~K. Jain, 3d touchless fingerprints:
  compatibility with legacy rolled images, in: Biometric {Consortium}
  {Conference}, 2006 {Biometrics} {Symposium}: {Special} {Session} on
  {Research} at the, 2006, pp. 1--6.

\bibitem{priesnitz2021overview}
J.~Priesnitz, C.~Rathgeb, N.~Buchmann, C.~Busch, An overview of touchless 2d
  fingerprint recognition, EURASIP Journal on Image and Video Processing 2021
  (2021) 25.

\bibitem{hiew2006automatic}
B.~Y. Hiew, A.~B.~J. Teoh, D.~C.~L. Ngo, Automatic {Digital} {Camera} {Based}
  {Fingerprint} {Image} {Preprocessing}, in: International {Conference} on
  {Computer} {Graphics}, {Imaging} and {Visualisation} ({CGIV}), 2006, pp.
  182--189.

\bibitem{sisodia2017conglomerate}
D.~S. Sisodia, T.~Vandana, M.~Choudhary, A conglomerate technique for finger
  print recognition using phone camera captured images, in: {International}
  {Conference} on {Power}, {Control}, {Signals} and {Instrumentation}
  {Engineering} ({ICPCSI}), 2017, pp. 2740--2746.

\bibitem{wang2016preprocessing}
K.~Wang, H.~Cui, Y.~Cao, X.~Xing, R.~Zhang, A preprocessing algorithm for
  touchless fingerprint images, in: Biometric Recognition, 2016, pp. 224--234.

\bibitem{malhotra_fingerphoto_2017}
A.~Malhotra, A.~Sankaran, A.~Mittal, M.~Vatsa, R.~Singh, Chapter 6 -
  fingerphoto authentication using smartphone camera captured under varying
  environmental conditions, in: Human Recognition in Unconstrained
  Environments, 2017, pp. 119 -- 144.

\bibitem{raghavendra2013scaling-robust}
R.~{Raghavendra}, C.~{Busch}, B.~{Yang}, Scaling-robust fingerprint
  verification with smartphone camera in real-life scenarios, in: Sixth
  International Conference on Biometrics: Theory, Applications and Systems
  (BTAS), 2013, pp. 1--8.

\bibitem{stein2012fingerphoto}
C.~{Stein}, C.~{Nickel}, C.~{Busch}, Fingerphoto recognition with smartphone
  cameras, in: International Conference of Biometrics Special Interest Group
  (BIOSIG), 2012, pp. 1--12.

\bibitem{priesnitz2020touchless}
J.~Priesnitz, C.~Rathgeb, N.~Buchmann, C.~Busch, Touchless fingerprint sample
  quality: Prerequisites for the applicability of nfiq2. 0, in: International
  Conference of the Biometrics Special Interest Group (BIOSIG), 2020, pp. 1--5.

\bibitem{fitzpatrick1988concept}
T.~B. Fitzpatrick, {The Validity and Practicality of Sun-Reactive Skin Types I
  Through VI}, Archives of Dermatology 124~(6) (1988) 869--871.

\bibitem{ISO19794-4}
ISO, {ISO/IEC 19794-4:2011}: {Information technology -- Biometric data
  interchange formats -- Part 4: Finger image data}, Standard, International
  Organization for Standardization (2011).

\bibitem{ISO19795-1}
I.~ISO, Iec 19795-1: Information technology--biometric performance testing and
  reporting-part 1: Principles and framework, ISO/IEC, Editor 1~(3) (2006) 5.

\bibitem{furman_contactless_2017}
S.~M. Furman, B.~C. Stanton, M.~F. Theofanos, J.~M. Libert, J.~D. Grantham,
  \href{https://nvlpubs.nist.gov/nistpubs/ir/2017/NIST.IR.8171.pdf}{Contactless
  fingerprint devices usability test}, Tech. Rep. NIST IR 8171, National
  Institute of Standards and Technology, Gaithersburg, MD (Mar. 2017).
\newblock \href {https://doi.org/10.6028/NIST.IR.8171}
  {\path{doi:10.6028/NIST.IR.8171}}.
\newline\urlprefix\url{https://nvlpubs.nist.gov/nistpubs/ir/2017/NIST.IR.8171.pdf}

\bibitem{porst_fragebogen_2014}
R.~Porst, Fragebogen: ein {Arbeitsbuch}, 4th Edition, Studienskripten zur
  {Soziologie}, Springer VS, Wiesbaden, 2014, oCLC: 870294421.

\bibitem{rohrmann_empirische_1978}
B.~Rohrmann, Empirische {Studien} zur {Entwicklung} von {Antwortskalen} für
  die sozialwissenschaftliche {Forschung}, Zeitschrift für Sozialpsychologie
  9~(3) (1978) 222--245.

\bibitem{Tang-FingerNet-2017}
Y.~Tang, F.~Gao, J.~Feng, Y.~Liu, {FingerNet}: An unified deep network for
  fingerprint minutiae extraction, in: International Joint Conference on
  Biometrics ({IJCB}), IEEE, 2017, pp. 108--116.

\bibitem{Vazan-SourceAFIS-2019}
R.~Va\v{z}an, {SourceAFIS} -- opensource fingerprint matcher,
  \url{https://sourceafis.machinezoo.com/}, last accessed: \today (2019).

\bibitem{mann_test_1947}
H.~B. Mann, D.~R. Whitney, \href{www.jstor.org/stable/2236101}{On a {Test} of
  {Whether} one of {Two} {Random} {Variables} is {Stochastically} {Larger} than
  the {Other}}, The Annals of Mathematical Statistics 18~(1) (1947) 50--60.
\newline\urlprefix\url{www.jstor.org/stable/2236101}

\bibitem{1263277}
J.~{Ortega-Garcia}, J.~{Fierrez-Aguilar}, D.~{Simon}, J.~{Gonzalez},
  M.~{Faundez-Zanuy}, V.~{Espinosa}, A.~{Satue}, I.~{Hernaez}, J.~. {Igarza},
  C.~{Vivaracho}, D.~{Escudero}, Q.~. {Moro}, Mcyt baseline corpus: a bimodal
  biometric database, IEE Proceedings - Vision, Image and Signal Processing
  150~(6) (2003) 395--401.

\bibitem{CAPPELLI20077}
R.~Cappelli, M.~Ferrara, A.~Franco, D.~Maltoni, Fingerprint verification
  competition 2006, Biometric Technology Today 15~(7-8) (2007) 7--9.

\bibitem{kumar_hong_2017}
A.~Kumar, \href{http://www4.comp.polyu.edu.hk/ csajaykr/fingerprint.htm}{The
  {Hong} {Kong} {Polytechnic} {University} {Contactless} {2D} to
  {Contact}-based {2D} {Fingerprint} {Images} {Database} {Version} 1.0} (2017).
\newline\urlprefix\url{http://www4.comp.polyu.edu.hk/ csajaykr/fingerprint.htm}

\bibitem{olsen_fingerprint_2015}
M.~A. Olsen, M.~Dusio, C.~Busch, Fingerprint skin moisture impact on biometric
  performance, in: 3rd {International} {Workshop} on {Biometrics} and
  {Forensics} ({IWBF} 2015), 2015, pp. 1--6.

\bibitem{oconnell_case_2020}
K.~A. O'Connell, C.~W. Enos, E.~Prodanovic, Case {Report}:
  {Handwashing}-{Induced} {Dermatitis} {During} the {COVID}-19 {Pandemic},
  American Family Physician 102~(6) (2020) 327--328.

\bibitem{tan_contact_2020}
S.~W. Tan, C.~C. Oh, Contact {Dermatitis} from {Hand} {Hygiene} {Practices} in
  the {COVID}-19 {Pandemic}, Annals of the Academy of Medicine, Singapore
  49~(9) (2020) 674--676.

\bibitem{otter_transmission_2016}
J.~A. Otter, C.~Donskey, S.~Yezli, S.~Douthwaite, S.~D. Goldenberg, D.~J.
  Weber, Transmission of {SARS} and {MERS} coronaviruses and influenza virus in
  healthcare settings: the possible role of dry surface contamination, Journal
  of Hospital Infection 92~(3) (2016) 235--250.

\bibitem{union2019implementing}
E.~Union, \href{http://data.europa.eu/eli/dec_impl/2019/329/oj}{Commission
  implementing decision (eu) 2019/329 of 25 february 2019 laying down the
  specifications for the quality, resolution and use of fingerprints and facial
  image for biometric verification and identification in the entry/exit system
  (ees)}, Official Journal of the European Union (2019) 18--28.
\newline\urlprefix\url{http://data.europa.eu/eli/dec_impl/2019/329/oj}

\bibitem{verifinger_neuro_2010}
S.~VeriFinger, Neuro {Technology}, VeriFinger, SDK Neuro Technology (2010).

\bibitem{vyas_collaborative_2019}
R.~Vyas, A.~Kumar, A {Collaborative} {Approach} using {Ridge}-{Valley}
  {Minutiae} for {More} {Accurate} {Contactless} {Fingerprint}
  {Identification}, arXiv:1909.06045 [cs, eess] (Sep. 2019).

\end{thebibliography}
\end{document}